\documentclass{article}

\usepackage[final,nonatbib]{neurips_2023}

\usepackage[utf8]{inputenc} 
\usepackage[T1]{fontenc}    
\usepackage{hyperref}       
\usepackage{url}            
\usepackage{booktabs}       
\usepackage{amsfonts}       
\usepackage{nicefrac}       
\usepackage{microtype}      
\usepackage{xcolor}         
\usepackage{tcolorbox}
\usepackage{amsmath,amssymb,amsfonts}
\usepackage[scientific-notation=true]{siunitx}
\usepackage{adjustbox}
\usepackage{array}
\newcolumntype{P}[1]{>{\centering\arraybackslash}p{#1}}
\usepackage{chemfig}
\usepackage{caption}  
\usepackage{threeparttable}
\usepackage{pifont}  
\usepackage{ltablex}
\usepackage{xcolor}
\newcolumntype{b}{X}
\newcolumntype{s}{>{\hsize=2.2\hsize}X}

\title{Cross-Modal Learning for Chemistry Property Prediction: Large Language Models Meet Graph Machine Learning}
\vspace{-3mm}
\author{
  Sakhinana Sagar Srinivas\thanks{Designed and programmed research, conducted experiments, analyzed results, and drafted manuscript}, \textbf{Venkataramana Runkana}\\
  TCS Research \\
  \texttt{sagar.sakhinana@tcs.com, venkat.runkana@tcs.com} 
}

\begin{document}

\maketitle

\vspace{-2mm}
\begin{abstract}
\vspace{-4mm}
In the field of chemistry, the objective is to create novel molecules with desired properties, facilitating accurate property predictions for applications such as material design and drug screening. However, existing graph deep learning methods face limitations that curb their expressive power. To address this, we explore the integration of vast molecular domain knowledge from Large Language Models (LLMs) with the complementary strengths of Graph Neural Networks (GNNs) to enhance performance in property prediction tasks. We introduce a Multi-Modal Fusion (MMF) framework that synergistically harnesses the analytical prowess of GNNs and the linguistic generative and predictive abilities of LLMs, thereby improving accuracy and robustness in predicting molecular properties. Our framework combines the effectiveness of GNNs in modeling graph-structured data with the zero-shot and few-shot learning capabilities of LLMs, enabling improved predictions while reducing the risk of overfitting. Furthermore, our approach effectively addresses distributional shifts, a common challenge in real-world applications, and showcases the efficacy of learning cross-modal representations, surpassing state-of-the-art baselines on benchmark datasets for property prediction tasks.
\end{abstract}

\vspace{-7mm}
\section{Introduction}
\vspace{-4mm}
Deep learning has great potential for efficiently predicting molecular properties compared to traditional methods, resulting in reduced computational complexity and costs. Recently, there has been a growing surge of interest in representing organic molecules as molecular graphs. Graph Neural Networks (GNNs) can learn patterns from these highly non-linear molecular topological structures, which find applications in various domains, including drug discovery\cite{vamathevan2019applications, vert2023will}, material design\cite{pan2022deep, sridhara2022generalized}, and environmental science\cite{zhong2021machine, tuia2023artificial}. Several state-of-the-art graph machine learning algorithms are available in the literature, including MPNN\cite{gilmer2017neural}, SchNet\cite{schutt2017schnet}, E(n)-GNN\cite{satorras2021n}, DimeNet++\cite{gasteiger2020directional}, and SphereNet\cite{liu2021spherical}. GNNs have been extensively studied for molecular property applications; however, they suffer from bottlenecks such as limited expressive power\cite{li2022expressive}, over-squashing\cite{di2023over}, and over-smoothing\cite{rusch2023survey} issues. In recent years, Large Language Models (LLMs)\cite{brown2020language, chowdhery2022palm, touvron2023llama} have revolutionized the field of natural language processing with improved performance in various tasks, generating human-like responses, facilitating complex logical reasoning, and demonstrating capabilities in multitasking and multi-modal learning. Zero-Shot Chain of Thought\cite{wei2022chain}(for brevity, Zero-Shot CoT) and Few-shot (In-Context) Learning\cite{brown2020language}(for brevity, Few-Shot ICL) are prompt engineering strategies for utilizing LLMs in specific linguistic tasks or related problem-solving scenarios. Zero-Shot CoT relies on task-specific instructions without demonstrations, requiring the language model to generalize from the implicit knowledge embedded within its parameters, which was learned from the training data, to generate the output. Conversely, Few-Shot ICL includes a few guiding demonstrations along with the instructions in the prompts, fostering contextual understanding and task-specific adaptation. Despite the progress in the application of foundational LLMs across various scientific disciplines, integrating LLMs with GNNs for the task of molecule property prediction remains an underexplored area. However, this presents an opportunity for innovative techniques that combine LLMs and GNNs to enhance property prediction applications. GNNs are effective at modeling the complex, graph-structured molecular data, capturing the structural and feature characteristics of graphs. Meanwhile, LLMs can encode molecular information implicitly within their parameters, owing to their training on extensive and diverse text corpora. LLMs can provide linguistic insights rich in domain-specific knowledge that serve as auxiliary information, bolstering property prediction tasks. The goal is to leverage the complementary strengths of both LLMs and GNNs to create a more robust and accurate predictive framework. In this study, we introduce a novel multi-modal fusion framework, $\texttt{MMF}$, that integrates the complementary analytical capabilities of GNNs and linguistic comprehension of LLMs within an end-to-end design to enhance the accuracy and robustness of molecular predictions. This framework achieves superior performance in chemical property prediction compared to state-of-the-art baselines, thereby reducing the risk of overfitting and potentially accelerating both the training and inference processes. Overall, this work introduces the subsequent contributions, outlined as follows:

\vspace{-3mm}
\begin{itemize}
\item  We propose a multi-faceted semantic fusion approach to obtain cross-modal embeddings, which combines Zero-shot LLMs prompting with graph neural networks (GNNs). We employ a five-step approach to generate cross-modal embeddings for molecular graphs: (a) First, we use custom CoT prompts, which include task-specific instructions to query LLMs in a zero-shot setting to generate technical descriptions on various aspects of the chemical SMILES representations, such as functional groups and chemical properties. (b) Next, we fine-tune small-scale language models (LMs) using the generated technical descriptions for domain-specific customization to compute context-aware token embeddings.
(c) Then, we employ a softmax attention pooling mechanism to compute text-level embeddings from the contextualized token embeddings to encapsulate the rich domain-specific knowledge in the generated textual descriptions. (d) Independently and in parallel, we utilize pre-existing GNNs --- specifically, Chebyshev Graph Convolution (CGC)\cite{defferrard2016convolutional, he2022convolutional} --- to interpret complex, graph-structured molecular data to compute molecular graph-level embeddings. (e) Finally, we use a cross-modal multi-head attention mechanism\cite{vaswani2017attention} to integrate the graph and text-level embeddings, offering a robust and efficient framework for generating semantically enriched cross-modal embeddings that effectively bridge structured and unstructured data.
\vspace{-1mm}
\item We propose ICL for the few-shot molecular property prediction task with LLMs. We leverage ICL to guide LLMs in predicting molecular properties without the necessity of explicit fine-tuning on labeled data. Utilizing context-augmented prompts --- which comprise task-specific instructions and demonstrations (input-output mappings, in SMILES notation for molecules and their properties) --- Our method queries LLMs to generate and transform predictions into a prediction embedding. It capitalizes on the implicit knowledge embedded within the pretrained parameters of LLMs, enabling it to make accurate predictions for new, unseen molecules, conditioned on the context-augmented prompt.
\vspace{-1mm}
\item We use the Mixture-of-Experts (MOE) method with a gating mechanism at the output layer for high-precision molecule property prediction. Two types of embeddings, cross-modal and prediction embeddings, are integrated into a unified embedding using a gating mechanism that dynamically allocates weights to each embedding based on their predictive performance. The framework's training objectives are twofold: first, to optimize the weight distribution of each embedding to accurately predict the ground-truth molecular properties; second, to fine-tune the embeddings based on this weight distribution. Overall, the framework aims to leverage the strengths of multiple learning strategies to achieve high-precision predictions for molecular properties.
\end{itemize} 

\begin{figure}[ht!]
\centering
\resizebox{0.99\linewidth}{!}{ 
\hspace*{0mm}\includegraphics[keepaspectratio,height=5.0cm,trim=0.0cm 3.0cm 0cm 1.0cm,clip]{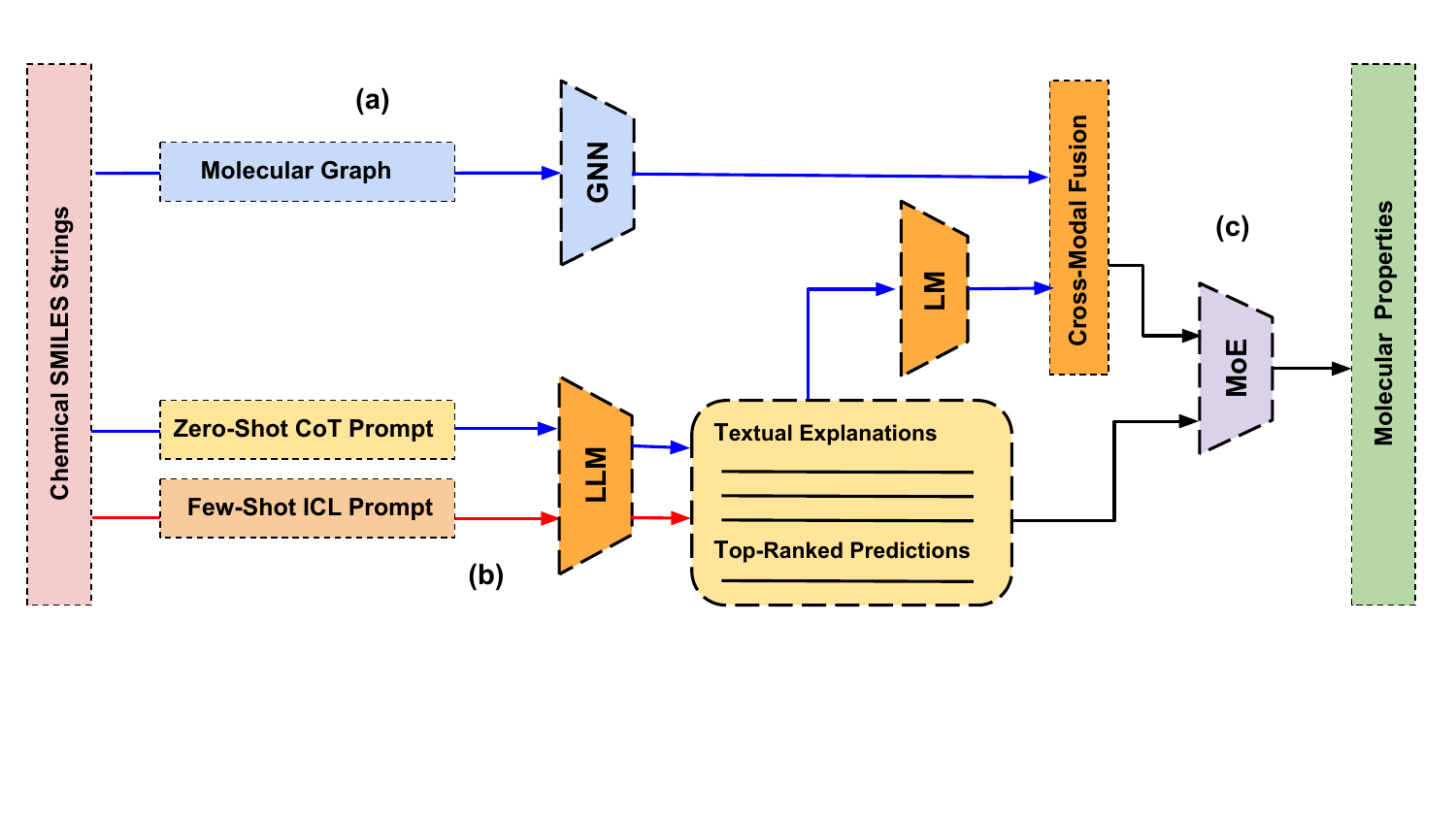} 
}
\vspace{-4mm}
\caption{Overview of \texttt{MMF} framework. Our framework leverages both the generative and predictive abilities of LLMs. The proposed molecular property prediction framework is a robust, efficient, and multi-step pipeline for predicting molecular properties with high precision. (a) Firstly, it introduces a multi-faceted semantic fusion strategy that leverages Zero-shot CoT prompting of LLMs approach alongside GNNs to generate semantically-aligned cross-modal embeddings for molecules, seamlessly integrating structured and unstructured data. (b) Secondly, the framework incorporates ICL, which taps into the inherent knowledge within pre-trained parameters of LLMs to make accurate predictions on new, unseen molecules, generating prediction embeddings guided by context-augmented prompts without the necessity for explicit fine-tuning on labeled data. (c) Lastly, it employs a MOE mechanism that integrates cross-modal and prediction embeddings through a gating mechanism at the output layer and optimizes the unified embeddings for downstream supervised regression tasks to achieve high-precision predictions. Overall, the cohesive framework aims to synergize multiple learning strategies to achieve unparalleled precision and efficiency in molecular property predictions. It is important to note that we do not customize LLMs through fine-tuning for task-specific adaptation. Instead, we access LLMs through LMaaS\cite{sun2022black} platforms via text-based API interaction. The three steps (a), (b), and (c) are illustrated with blue, red, and black arrow lines.}
\label{fig:figure1}
\vspace{-2mm}
\end{figure}

\vspace{-3mm}
Experiments were conducted using six publicly available molecule property prediction datasets to evaluate the performance of the \texttt{MMF} framework, with consistent results demonstrating its effectiveness in accurately predicting properties across all benchmark datasets. In summary, we present a cohesive and multifaceted framework that integrates advanced computational approaches and learning strategies to enhance precision and efficiency in molecular property predictions, potentially fostering advancements in molecular science and technology. The workflow of the proposed approach is illustrated in Figure \ref{fig:figure1}.

\vspace{-4mm}
\section{Proposed Method}
\vspace{-3mm}

\subsection{Task Formulation}
\vspace{-3mm} 
A molecular graph $\mathcal{G}$ consists of a set of nodes(atoms) $\mathcal{V}$ and edges(bonds) $\mathcal{E}$ with node feature($\mathbf{X}^{v} \in \mathbb{R}^{|\mathcal{V}| \times \mathbf{d}_{v}}$) and edge feature matrix(${\small\mathbf{X}^{e}} \in \mathbb{R}^{|\mathcal{E}| \times \mathbf{d}_{e}}$), where $\mathbf{d}_{v}$, $\mathbf{d}_{e}$ denote the dimensions of nodes and edge features, respectively. The adjacency matrix, $\mathcal{G}_{A} \in [0,1]^{|\mathcal{V}| \times |\mathcal{V}|}$, describes the graph structure, where $\mathcal{G}_{A}[v, u]=1$ if $(v,u) \in \mathcal{E}, u, v \in \mathcal{V}$ or else $\mathcal{G}_{A}[v, u]=0$. In the graph property prediction task, let $\mathcal{D}_L= (\mathcal{G}_L, \mathcal{P}_L)$ be the labeled dataset, comprising a set of graphs denoted as $\mathcal{G}_L$, with corresponding properties $\mathcal{P}_L$. A graph encoder, denoted by $f_{\gamma}(\mathcal{G}) \rightarrow \mathbf{h}_{g}$, is trained on labeled dataset $\mathcal{D}_L$ to obtain graph-level embeddings $\mathbf{h}_{g}$, where $\gamma$ represents the trainable parameters of graph encoder. A pre-trained LM encoder, denoted by $f^{\prime}_{\theta}(\mathcal{S}_{e}) \rightarrow \mathbf{h}_\text{text}$, is fine-tuned on technical descriptions($\mathcal{S}_{e}$) generated by zero-shot CoT prompting of LLMs on molecular graphs $\mathcal{G}$ from the labeled dataset $\mathcal{D}_L$ to compute text-level embedding $\mathbf{h}_\text{text}$. $\theta$ represents the trainable parameters of LM encoder. A few-shot ICL prompting of LLMs with a few input-output pairs($\mathcal{G}, p$) from the labeled dataset $\mathcal{D}_{L}$ to compute the predictive embedding, $\mathbf{h}_\text{ICL}$. The joint-optimization objective function for property prediction task is defined as minimizing the regression loss $\mathcal{L}_{\text{graph}}$ using a supervised-learning approach to predict the properties $\mathcal{P}_U$ of unlabeled graphs $\mathcal{G}_U$, described as follows,

\vspace{-5mm}
\resizebox{0.95\linewidth}{!}{
\begin{minipage}{\linewidth}
\begin{equation}
 \min_{\gamma, \theta, \omega} \mathcal{L}_{graph}\left(\mathcal{G}_{i}, \gamma, \theta, \omega\right)=\sum_{\left(\mathcal{G}_{i}, p_{i}\right) \in \mathcal{D}_{L}} \ell\bigg(g_\omega \big(\mathbf{h}_{g_i}, h_{\text{text}_i}, h_{\text{ICL}_i}\big), p_{i}\bigg)
 \end{equation}
 \end{minipage}
}

\vspace{-2mm}
where, the non-linear function $g_\omega(\cdot)$  linearly maps an input vector to a single output value, $\omega$ denote the trainable parameters of the non-linear function. $\ell(\cdot, \cdot)$ denotes the mean squared error loss.

\vspace{-5mm}

\vspace{-6mm}
\subsection{Graph Chebyshev Convolution}
\label{gcc}
\vspace{-3mm}
Graph Convolutional Networks (GCNs) are designed to learn from graph data. They can be categorized into spatial and spectral approaches. The spatial GCN analyzes node neighborhoods. The spectral GCN, grounded in the spectral graph theory, uses the eigenvalues and eigenvectors of the graph's Laplacian matrix for convolutions. The spectral convolution\cite{bruna2013spectral} is computationally expensive and inherently non-local. Graph Chebyshev convolution(CGC, \cite{defferrard2016convolutional, he2022convolutional}) is a scalable alternative to spectral convolution that offers locality in capturing local-graph-based features, flexibility in approximating spectral properties, and scalability through recursive Laplacian computation. CGC operator utilizes Chebyshev polynomials to approximate spectral graph convolution, enabling the application of convolutional filters on graph-structured data by approximating the graph Laplacian with Chebyshev polynomials. Chebyshev polynomials are obtained from the normalized Laplacian matrix of a graph, $\hat{L} = \hat{D}^{-1/2} \hat{\mathcal{G}}_A \hat{D}^{-1/2}$, where $\hat{\mathcal{G}}_A$ is the normalized adjacency matrix and $\hat{D}$ is the diagonal degree matrix. The Chebyshev approximation approximates the graph Laplacian using Chebyshev polynomials, designated as $T_k(\hat{L})$, calculated through a recurrence relation described as,

\vspace{-1mm}
\resizebox{0.875\linewidth}{!}{
\begin{minipage}{\linewidth}
\[
\text{T}_k(\hat{L}) = 
\begin{cases}
I, & \text{if } k = 0 \\
\hat{L}, & \text{if } k = 1 \\
2\hat{L}\text{T}_{k-1}(\hat{L}) - \text{T}_{k-2}(\hat{L}), & \text{otherwise}
\end{cases}
\]
\end{minipage}
}

where k refers to the degree of the Chebyshev polynomial and  $I$ is identity matrix. Given an input graph-based node and edge feature matrix, \resizebox{.03\textwidth}{!}{$\mathbf{X}^{v}$} and \resizebox{.03\textwidth}{!}{$\mathbf{X}^{e}$}, respectively. The Chebyshev graph convolution operation is defined as follows,

\vspace{-2mm}
\resizebox{0.875\linewidth}{!}{
\begin{minipage}{\linewidth}
\begin{equation}
\text{h}_{CGC} = \sigma\left(\sum_{k=0}^{K-1} \text{T}_k(\hat{L}) (\text{W}_{0}\mathbf{X}^{v} + \text{W}_{1} \mathbf{X}^{e})\Theta_k\right)
\end{equation}
\end{minipage}
}

\vspace{-1mm}
where $\text{W}_{0} \in \mathbb{R}^{d \times d_{v}}$, $\text{W}_{1} \in \mathbb{R}^{d \times d_{e}}$ denote the trainable weight matrices and $\sigma(\cdot)$ denotes the non-linear sigmoid activation function. $\Theta_k \in \mathbb{R}^{\hspace{0.5mm}d \times d}$ is the weight or parameter matrix associated with the $k$th-order Chebyshev polynomial. In summary, the layerwise differentiable neural operator maps discrete graphs to a node-level embedding matrix, $\text{h}_{\text{GCC}} \in \mathbb{R}^{\hspace{0.25mm}n \times d}$, that maximally captures both the topology and the feature information embedded within the graphs. We perform global-graph pooling using the Set2Set algorithm\cite{vinyals2015order} to aggregate and summarize the nodes feature information to obtain a graph-level embedding($\mathbf{h}_{g} \in \mathbb{R}^{d}$) to encapsulate and preserve the overall graph characteristics.

\vspace{-5mm}
\subsection{Language Models} 
\label{lms}
\vspace{-3mm}
In recent years, the introduction of pre-trained large language models (LLMs), such as ChatGPT\cite{brown2020language}, Google's PaLM\cite{chowdhery2022palm}, and Meta's LLaMA\cite{touvron2023llama}, has had a transformative impact in the domain of language modeling, enhancing performance and capabilities across a wide range of NLP tasks and applications. The LLMs adopt a `pre-train, prompt, and predict' approach, attaining vast linguistic understanding through pre-training and generating human-like responses with tailored prompts\cite{wei2022chain, yao2023tree}. Smaller pre-trained language models (LMs), such as BERT\cite{devlin2018bert} and DeBERTa\cite{he2020deberta}, lack the sophisticated logical reasoning abilities of LLMs. However, they offer advantages such as access to logits or token embeddings, which aid in explainability for downstream applications utilizing the pre-trained LM models. In addition, these small-scale LMs can be fine-tuned with labeled data in an affordable manner for domain-specific customization. However, LLMs are resource-intensive to fine-tune with labeled data for task adaptation due to their high model complexity, making them less accessible to low-budget research labs. Additionally, the black-box nature of these large language models limits interpretability in downstream applications by not providing access to latent token embeddings or logits. To address these challenges, Language Modeling as a Service (LMaaS\cite{sun2022black}) allows text-based API access to LLMs avoiding the high computational costs of domain-specific customization through fine-tuning. While it may seem intuitive to use LLMs for interpreting chemical SMILES strings, the study of their effectiveness in predicting molecular properties is still in its early stages. LLMs have been proven to be effective at zero-shot learning and (in-context) few-shot learning across a diverse set of tasks in the field of NLP. We refer to the method of conditioning the language model as `prompting'. Prompts with explicit conditioning based on task-specific instructions and a few demonstrations are termed as `few-shot prompts', while those that rely solely on task-specific instructions are referred to as `zero-shot prompts'. In our molecular property prediction task, we focus on the development and exploration of various hand-crafted prompt engineering strategies. These include the zero-shot chain-of-thought (Zero-Shot CoT) and few-shot (in-context) learning (Few-Shot ICL), which aid in conditioning the LLMs to adapt to new tasks either through instructions that describe the task (zero-shot) without prior labeled data or through task-specific instructions accompanied by demonstrations(input-output pairs, i.e., few-shot). In our framework, we utilize LLMs in both zero-shot and few-shot learning scenarios, each serving a different purpose. (a) Our approach employs customized CoT prompts to query general-purpose LLMs in a zero-shot setting, generating textual descriptions that encapsulate various aspects of organic molecules, including molecular structure, physical properties, applications, and more. We fine-tune small-scale pre-trained LMs using these technical descriptions, then compute text-level embeddings using a softmax attention pooling mechanism to encapsulate the specialized knowledge in the technical descriptions, which is critical for downstream property prediction tasks. (b) We use (in-context) few-shot prompting of LLMs for molecular property prediction by leveraging input-output mappings (chemical SMILES strings --- molecular properties pairs) in the context-augmented prompts. The ICL approach exploits the inherent knowledge of LLMs, conditioning on the augmented prompt to accurately predict molecular properties for new, unseen molecules, without the need for explicit fine-tuning.

\vspace{-2mm}
\paragraph{Evaluation LLMs \& LMs:}
\vspace{-2mm}
In our work, we employ three representative LLMs: text-davinci-003, ChatGPT, and BARD. Table \ref{model} summarizes the main characteristics of the LLMs. Text-davinci-003, a GPT-3 model from OpenAI, excels in numerous natural language tasks, particularly zero-shot and few-shot instruction-following tasks. GPT-3.5-turbo, a refined variant within the GPT-3.5 model family, is widely recognized for its exceptional performance and cost-effectiveness. Meanwhile, Google's BARD\footnote{https://bard.google.com}, a recently updated chatbot featuring a new large language model (LLM) known as PaLM 2\cite{anil2023palm}, stands out due to its significantly larger parameter count and expanded vocabulary size compared to the models in the GPT-3.5 family. In the Google Bard and GPT model family, two parameters --- Top-p (also known as nucleus sampling or probabilistic sampling) and temperature --- are crucial for controlling text generation. Specifically, Top-p governs the diversity of generated text, while the temperature parameter influences the randomness of the language models during text generation. The Top-p parameter sets a probability threshold for including a token in the generated sentence, serving as a sampling method that prevents the language model from generating tokens that are either too rare or too common. Conversely, the temperature parameter modulates the randomness of the generated text: a higher temperature leads to more random text, whereas a lower temperature results in more deterministic text. In our experiments, we set the Top-p and temperature parameters to one and zero, respectively, to retrieve factual and accurate textual outputs. Additionally, we utilized a pre-trained small-scale language model known as DeBERTa\footnote{For more information, refer to the DeBERTa model documentation available at \url{https://huggingface.co/docs/transformers/index}.}\cite{he2020deberta}. In a zero-shot setting, LLMs generate technical descriptions about chemical SMILES strings, while small-scale language models encode rich knowledge in textual descriptions for task-specific customization, enhancing the performance of the framework in property prediction tasks. In contrast, we employ few-shot prompting of LLMs to predict molecular properties.

\vspace{-6mm}
\begin{table}[ht!]
	\centering
	\small
	\caption{Specifications of LLMs and LMs: \emph{Enterprise} refers to the technology organization that developed the language model; \emph{Charges} indicates the cost associated with the use of 1K tokens; \emph{Last Update Date} denotes that the LLM's knowledge base is limited to information available up until that date.}
	\vspace{-3mm}
	\begin{threeparttable}
	\hspace{-5mm}\begin{tabular}{c|c|c|c|c}
            \toprule
            \textbf{Model} &\textbf{Enterprise} &\textbf{Charges}  &\textbf{Last Update Date} &\textbf{Vocabulary size}\\
            \hline
            text-davinci-003 &Open-AI &0.02\$ &Sep. 2021 &175B  \\
            ChatGPT &Open-AI &0.002\$ &Jun. 2021 & 175B \\
            BARD & Google &Free & Undisclosed & 1,560B \\
            \hline
            DeBERTa & Hugging Face & Free & N/A & 50M \\
            \hline
	\end{tabular}
	\end{threeparttable}
	\vspace{-0.2cm}
	\label{model}
\end{table}

\vspace{-4mm}
\paragraph{Zero-short LLM Prompting:}\label{zeroshot}  We access LLMs through LMaaS platforms\cite{sun2022black} via text-based API interactions. We utilize a standardized chain-of-thoughts (CoT) prompt template to query LLMs in a zero-shot setting, aiming to glean linguistic insights into the specialized knowledge of chemical SMILES strings, including their structure and physical properties, among others. The generated technical descriptions serve as auxiliary information for downstream applications. The prompt template guides the LLMs through a sequence of open-ended queries to acquire specific information about a given organic molecule. The custom CoT prompt format is as follows:

\vspace{-2mm}
\begin{tcolorbox}[colframe=black, colback=white, boxrule=0.5mm, width=\textwidth, sharp corners, left=2mm, right=2mm]
\vspace{-1mm}
\textbf{Prompt 1:} What is the molecular structure of this \textbf{chemical SMILES strings} Could you describe its atoms, bonds, functional groups, and overall arrangement?
\textbf{Prompt 2:} What are the physical properties of this molecule such as its boiling point, melting point, and density?
\textbf{Prompt 3:} What is the solubility behavior of this molecule? In which solvents does it dissolve and which does it not?
\textbf{Prompt 4:} What is the chemical reactivity of this molecule? How does it interact with various reagents?
\textbf{Prompt 5:} Are there any common reactions that this molecule is known to undergo? Could you describe them?
\textbf{Prompt 6:} What is the mechanism of these reactions? Could you describe the various steps involved?
\textbf{Prompt 7:} Does this molecule exhibit any unique optical, electrical, or magnetic properties?
\textbf{Prompt 8:} Is this molecule chiral? If yes, how does its chirality influence its behavior or properties?
\textbf{Prompt 9:} Does this molecule form part of any important biological processes or pathways?
\textbf{Prompt 10:} Is this molecule synthesized industrially or in the laboratory? If yes, could you explain the process?
\textbf{Prompt 11:} Is this molecule found naturally? If yes, in what sources is it most commonly found?
\textbf{Prompt 12:} Are there any notable uses or applications for this molecule in medicine, industry, or other fields?
\textbf{Prompt 13:} What safety measures should be taken when handling this molecule?
\textbf{Prompt 14:} Are there any environmental impacts associated with the production, use, or disposal of this molecule?
\vspace{-1mm}
\end{tcolorbox}

\vspace{-1mm}
Querying the LLM produces detailed technical descriptions of chemical SMILES strings and their properties.

\vspace{-1mm}
\begin{tcolorbox}[colframe=black, colback=white, boxrule=0.5mm, width=\textwidth, sharp corners, left=2mm, right=2mm]
\centering
\vspace{-2mm}
(\textbf{LLMs Response}) [Textual descriptions]
\vspace{-2mm}
\end{tcolorbox}

\vspace{-1mm}
In the next section, we will discuss how to integrate these textual descriptions as additional features to aid in the improvement and fine-tuning of downstream LMs and subsequent applications.

\vspace{-2mm}
\paragraph{Fine-Tuning LMs and Domain-Specific Customization:}
\vspace{-2mm}
In our approach, we utilize a small-scale pre-trained language model (LM) to encode the text outputs generated by a larger language model (LLM). During fine-tuning, the smaller LM extracts informative features from the generated descriptions for task-specific customization. The small-scale LM serves as an intermediate layer between LLMs and downstream prediction layers. We fine-tune small-scale LMs (referred to as $\textrm{LM}_{\textrm{expl}}$) to process technical descriptions generated by LLMs for the property prediction task. We input text sequences from LLMs (denoted as $\mathcal{S}_{e}$) into the $\textrm{LM}_{\textrm{expl}}$ model to compute context-aware token embeddings. These embeddings capture the contextual information and semantic relationships between the words or phrases described as follows,

\vspace{-3mm}
\resizebox{0.965\linewidth}{!}{
\hspace{0mm}\begin{minipage}{\linewidth}
\begin{equation}
h_{\textrm{expl}} = \textrm{LM}_\textrm{expl}(\mathcal{S}_{e});
\end{equation}
\end{minipage}
} 

\vspace{-1mm}
where the contextualized embeddings $h_{\textrm{expl}} \in \mathbb{R}^{m \times d}$, where $m$ represents the number of tokens in the input sequence $\mathcal{S}_{e}$ and $d$ is the token embedding dimension. To encode the textual explanations into a fixed-length vector, we apply a softmax attention pooling mechanism to calculate a weighted sum of the token embeddings. This results in a comprehensive representation of the entire textual descriptions, computed as follows,

\vspace{-3mm}
\resizebox{0.965\linewidth}{!}{
\hspace{0mm}\begin{minipage}{\linewidth}
\begin{equation}
\alpha_i = \mbox{softmax}(q_i); \hspace{2mm} q_i = \mathbf{u}^T h^{(i)}_{\textrm{expl}}
\end{equation}
\end{minipage}
} 

\vspace{-2mm}
\resizebox{0.935\linewidth}{!}{
\hspace{0mm}\begin{minipage}{\linewidth}
\begin{equation}
h_{\text{text}} = \sum_{i=0}^m{\alpha_i h^{(i)}_{\textrm{expl}}}
\end{equation}
\end{minipage}
} 

\vspace{-2mm}
where $\mathbf{u}$ is a trainable parameter and $\alpha$ is the attention coefficient. The text-level embedding $h_{\text{text}} \in \mathbb{R}^{d}$ encapsulates the extracted domain-specific knowledge from foundational LLMs on chemical SMILES strings. $h_{\text{text}}$ enhances explainability by unpacking the black-box nature of LLMs by utilizing the generated descriptions from LLMs on chemical SMILES strings.

\vspace{-4mm}
\paragraph{Few-Shot LLM Prompting:}\label{fewshot} In-Context Learning (ICL) enables LLMs to adapt to new tasks without the need for explicit, gradient-based fine-tuning\cite{brown2020language} on labeled data. This approach allows LLMs to learn through analogy, utilizing just a few input-output pairs specific to the downstream task. ICL leverages the implicit knowledge embedded in pre-trained LLM parameters to adapt to new tasks through task-specific demonstrations, thereby avoiding the need to repurpose LLMs with parameter updates. The context-augmented prompt provides task-specific instructions and demonstrations (input-output mappings), enabling LLMs to generate outputs conditioned on the prompt for improved generalization performance. In the case of molecular graph property prediction tasks, ICL involves constructing a context-augmented prompt using a few input-output pairs $(\mathcal{G}_i, \mathcal{P}_i)$ sampled from the training data and the task-specific instruction is related to the query SMILES representation. 
At inference time, on test input $\mathcal{G}_{\text{test}}$, ICL generates the output based on the conditional probability distribution, $\mathcal{P}_{\text{test}} \sim \mathbf{P}(\mathcal{P}_{\text{test}} \mid (\mathcal{G}_{\text{train}}, \mathcal{P}_{\text{train}}), \mathcal{G}_{\text{test}})$, where $\sim$ denotes the decoding strategy. To examine how the quality and quantity of ICL demonstrations impact the performance of property prediction tasks, we investigate two distinct ICL sampling strategies.  We explore two distinct ICL sampling strategies: ``Random” and ``Scaffold”. The quality of demonstrations is determined by the sampling strategies used to identify the top-$K$ chemical SMILES representations that are most similar to the query SMILES representation. To investigate the impact of the quantity of ICL demonstrations on performance, we optimize the number of ICL demonstrations ($K$) for each query SMILES representation. In the random strategy, we randomly sample $K$ input-output pairs from the training data. In contrast, the Scaffold strategy employs Tanimoto similarity \cite{tanimoto1958elementary} based on Morgan fingerprints \cite{morgan1965generation} with a radius of 2, to identify the top-$K$ most similar chemical SMILES representations to a given query SMILES representation within the training data. We employ two sampling strategies to construct an augmented prompt for analyzing the effectiveness of ICL demonstrations on property prediction tasks. In summary, our goal is to task LLMs with a contextual prompt. This prompt comprises a list of input-output pairs, where the input represents an organic molecule in SMILES notation, and the output denotes its molecular properties along with task-specific instructions. The instruction in the context-augmented prompt directs LLMs to predict the multiple molecular properties of the query SMILES representation. This task will demonstrate the LLM's ability to predict these properties based on its inherent knowledge, simply by conditioning on the prompt, without any parameter updates. This approach stands in contrast to supervised learning, where parameter updates are performed to fine-tune models based on labeled data, enabling them to predict the properties of new, unseen molecules. For each query SMILES representation, the PLLMs generate a \( c \)-dimensional vector \( h_{\text{pred}_i} \in \mathbb{R}^c \), where \( c \) signifies the dimension associated with the multiple properties to be predicted. This vector is linearly encoded into a high-dimensional space to produce a prediction embedding \( h_{\text{ICL}_i} \in \mathbb{R}^d \), which encapsulates the LLMs' predictions. Here, \( d \) represents the embedding dimension and \( c \ll d \). An example of an ICL prompt is as follows,

\vspace{-3mm}
\begin{tcolorbox}[colframe=black, colback=white, boxrule=0.5mm, width=\textwidth, sharp corners, left=2mm, right=2mm]
\centering
\vspace{-2mm}
Below are the input-output examples (SMILES strings-molecular properties pairs) for property prediction task. Predict the molecular properties for the query SMILES strings.
\vspace{-2mm}
\end{tcolorbox}

\vspace{-2mm}
In the subsequent sections, we will discuss the cross-modal attention layer and an output layer.

\vspace{-5mm}
\subsection{Cross-modal Attention Layer}
\vspace{-4mm}
We compute the cross-modal embedding $h^{f}_{i}$ through a multi-head attention mechanism\cite{vaswani2017attention} that integrates the graph-level embeddings $h_{g_i}$ and text-level embeddings $h_{\text{text}_i}$. The multi-head attention(MHA) mechanism offers a robust framework for integrating cross-domain embeddings, facilitating  multi-faceted analysis in the context of interpreting correlations between molecular structures and properties. (a) \textbf{Richer Representation:} The multi-head attention mechanism enables the computation of a richer, multi-faceted representation for each molecule. It allows each attention head to focus on different facets, such as chemical bonds or functional groups in the molecular structure. Meanwhile, other heads attend to corresponding textual descriptions, like reactivity or the physicochemical properties of the observed functional group.  (b) \textbf{Semantic Matching:} Additionally, the multi-head mechanism facilitates semantic matching, as individual heads can specialize in aligning various semantic aspects, such as specific functional groups in the molecular graph with their corresponding textual descriptions. For example, one attention head might specialize in aligning aromatic rings in the molecular structure with textual descriptions related to aromaticity. This allows the model to effectively integrate different types of molecular information, providing a contextual and comprehensive view of both text and graph data. (c) \textbf{Enhanced Context Sensitivity:} Finally, the multi-head mechanism enhances context sensitivity. Molecules often behave differently depending on their context. For instance, a molecule's reactivity can change based on its surrounding environment, and this information might be captured in textual descriptions. Multi-head attention allows the model to be sensitive to this context by considering both the graph-level embeddings and the text-level descriptions in tandem. This could be particularly beneficial for complex tasks. In summary, MHA mechanism for computing cross-modal embeddings facilitates a rich, nuanced representation of molecules by allowing parallel focus on various facets of data, including molecular graph structures and textual descriptions. It enhances semantic matching by aligning different semantic features from both text and graph data, and improves context sensitivity, enabling the model to understand and adapt to the dynamic behaviors of molecules in different contexts. Moreover, the MHA mechanism improves computational efficiency through parallel processing across heads, and its modular nature makes it easy to adapt and extend. It also offers benefits such as increased model capacity, and better generalization to unseen data. Overall, multi-head attention provides a robust and versatile framework for seamlessly fusing information across different modalities. We  compute the Query, Key, Value projections for graph-level embeddings for each head h as follows:

\vspace{-4mm}
\resizebox{0.945\linewidth}{!}{
\hspace{0mm}\begin{minipage}{\linewidth}
\begin{align}
Q^h_{g_i} &= h_{g_i} W^h_{Q_g}; K^h_{g_i} = h_{g_i} W^h_{K_g}; V^h_{g_i} = h_{g_i} W^h_{V_g}
\end{align}
\end{minipage}
} 

\vspace{-1mm}
Similarly, the Query, Key, Value projections for text-level embeddings for each head h:

\vspace{-3mm}
\resizebox{0.945\linewidth}{!}{
\hspace{0mm}\begin{minipage}{\linewidth}
\begin{align}
Q^h_{\text{text}_i} &= h_{\text{text}_i} W^h_{Q_{\text{text}}}; K^h_{\text{text}_i} = h_{\text{text}_i} W^h_{K_{\text{text}}}; V^h_{\text{text}_i} = h_{\text{text}_i} W^k_{V_{\text{text}}}
\end{align}
\end{minipage}
} 

We concatenate the keys and values from both graph-level and text-level embeddings, which provides a powerful way to integrate information from both modalities into a unified, rich representation. 

\vspace{-3mm}
\resizebox{0.945\linewidth}{!}{
\hspace{0mm}\begin{minipage}{\linewidth}
\begin{align}
K^h_{\text{concat}_i} &= [K^h_{g_i}, K^h_{\text{text}_i}]; V^h_{\text{concat}_i} = [V^h_{g_i}, V^h_{\text{text}_i}]
\end{align}
\end{minipage}
} 

We perform softmax attention to integrate complementary information from the two modalities, focus on contextually relevant information, and semantically align them through attention mechanism. The Softmax function is applied over the keys for each query.

\vspace{-4mm}
\resizebox{0.935\linewidth}{!}{
\hspace{0mm}\begin{minipage}{\linewidth}
\begin{align}
A^h_i &= \text{Softmax}\left(\frac{(Q^h_{g_i} + Q^h_{\text{text}_i}) {K^h_{\text{concat}_i}}^T}{\sqrt{d_h}}\right)
\end{align}
\end{minipage}
} 

\vspace{-2mm}
Each head outputs a new vector representation that highlights the most relevant features in the input embeddings (both graph and text-level), according to the attention mechanism for that specific head, which is tailored to the specific aspects or relationships within the data.

\vspace{-4mm}
\resizebox{0.945\linewidth}{!}{
\hspace{0mm}\begin{minipage}{\linewidth}
\begin{align}
O^h_i &= A^h_i V^h_{\text{concat}_i}
\end{align}
\end{minipage}
} 

\vspace{-1mm}
Finally, all the head-specific outputs are concatenated and linearly transformed to create the final cross-modal representation as follows,

\vspace{-4mm}
\resizebox{0.945\linewidth}{!}{
\hspace{0mm}\begin{minipage}{\linewidth}
\begin{align}
O_{\text{concat}_i} &= [O^1_i, O^2_i, \ldots, O^H_i] \\
h^f_i &= O_{\text{concat}_i} W_{O}
\end{align}
\end{minipage}
} 

where $W^h_{Q_g}$, $W^h_{K_g}$, $W^h_{V_g}$, $Q^h_{\text{text}_i}$,  $W^h_{Q_{\text{text}}}$,  $W^h_{K_{\text{text}}}$, $W_{O}$ are the learnable weight matrices. $d_h$ is the dimensionality of the key/query/value for each head, and $H$ is the number of heads. 

\vspace{-6mm}
\subsection{Output Layer} 
\vspace{-3mm}
In this framework, we utilize the mixture-of-experts (MOE) technique with a gating mechanism for the output layer. In the context of this framework, the MOE mechanism is a competitive game where each embedding --- either cross-modal embeddings obtained from the multi-head attention mechanism (integrating zero-shot LLM prompting \& GNNs outputs) or prediction embeddings  from few-shot LLM prompting --- aims to maximize its contribution to the final prediction. Unlike a cooperative game, each embedding aims to outperform the others for a larger weight from the gating mechanism. The gating mechanism allocates weights based on individual performance, creating a competitive landscape where more accurate embeddings gain greater influence. This competition can drive the framework towards a globally optimal solution. The embeddings are combined by the gating mechanism, which allocates input-dependent weights to calculate a weighted sum of embeddings. Training in this framework aims to: a) determine the optimal weight distribution for precise predictions of ground-truth molecular properties, and b) optimize the embeddings jointly according to the weight distribution specified by the gating mechanism. The unified embeddings are obtained by merging embeddings using input-dependent weights allocated by the gating mechanism as follows,

\vspace{-7mm}
\resizebox{0.985\linewidth}{!}{
\begin{minipage}{\linewidth}
\begin{align}
g &= \sigma \big( f_s(h^f_i) + f_g(h_{\text{ICL}_{i}}) \big) \\
\mathbf{h}_{u}  &= \sigma \big( g(h^f_i) + (1-g)(h_{\text{ICL}_{i}}) \big)
\end{align}
\end{minipage}
} 

\vspace{-1mm}
where \resizebox{.025\textwidth}{!}{$f_s$} and \resizebox{.025\textwidth}{!}{$f_g$} are linear operators and $\sigma$ is the non-linear sigmoid operation. Finally, we use a linear operator to transform $h_{u}$ to predict the molecular properties of each graph.

\vspace{-6mm}
\section{Experiments and Results}

\vspace{-5mm}
\subsection{Datasets and Experimental Setup}
\vspace{-3mm}
The QM8\cite{ramakrishnan2015electronic} and QM9\cite{ramakrishnan2014quantum} are two large datasets of quantum chemical properties for low-tree width organic molecules, which serve as benchmarks and training data for ML models in the field of quantum chemistry property prediction task. The QM8 dataset comprises 21,786 molecules containing up to 8 heavy atoms, namely C, O, N, or F. It provides properties such as electronic excitation energies, oscillator strengths, and ionization potentials. In comparison, the QM9 dataset is larger, consisting of 133,885 molecules with up to 9 heavy atoms, including properties like atomization energy, HOMO/LUMO gap, dipole moment, and polarizability. Both datasets are split into training, validation, and test sets. The training set is used to fit the model parameters, the validation set is used to select the best hyperparameters, and the test set is used to evaluate the model's generalization performance. The quantum properties were standardized to have zero mean and unit variance. The predictions were then re-normalized to the original scale to calculate the error metric. In this work, our framework integrates both large language models (LLMs) and smaller language models (LMs). We focus on employing zero-shot CoT and few-shot ICL learning techniques to prompt LLMs for the molecule prediction task without the need for retraining or finetuning. We achieve this by using a LMaaS platform\cite{sun2022black} to access frozen trainable parameters of LLMs through text-based API interactions. The hyperparameters of this framework were set to a batch size of 32, 50 epochs for training, and a hidden or embedding dimension of 128. We conducted our experiments using the following four large language models (LLMs): GPT-4.0, GPT-3.5-turbo, GPT-3.0-text-davinci-003, and Google Bard. We did not fine-tune the hyperparameters of our framework for each LLM. Instead, we utilized the same hyperparameters across all LLMs. This shows that our framework is general and easy to use, and that it can utilize any off-the-shelf LLMs. To optimize the use of computational resources, we utilized 8 V100 GPUs, each equipped with 8 GB of GPU memory, for the training of deep learning models built upon the PyTorch framework. The LLMs have a context length limitation with a maximum sequence length of 4096 tokens for GPT models and 4000 tokens for Google Bard. The Adam optimizer\cite{kingma2014adam} was used to train the framework, starting with a learning rate of $1e^{-3}$. A learning rate decay scheduler was employed to reduce the learning rate by half if the validation loss did not improve for 7 epochs, and early stopping was implemented to prevent overfitting on the training set. For few-shot learning, we utilized the scaffold technique with a hyperparameter $K$ set to 16 for sampling demonstrations to construct context-augmented prompts. The framework's performance was evaluated using the MAE metric, and the results were presented on the test datasets. Three independent experiments were conducted, and we report the average.

\vspace{-4mm}
\subsection{Results}
\vspace{-3mm}
Table \ref{table:qm8} compares the performance of the \texttt{MMF W/GPT-4} framework to baseline algorithms on the QM8 dataset. The results are reported on both the validation and test datasets. The primary objective is to predict 16 distinct properties of the electronic spectra and the energy per molecular graph. We report the average prediction error across all properties as a single value per molecular graph, maintaining consistency with earlier studies\cite{liao2019lanczosnet} for a fair comparison with the baseline algorithms. We report the baseline results from a previous study\cite{liao2019lanczosnet}. Our framework performance is compared against several baseline algorithms, namely GCN-FP \cite{duvenaud2015convolutional}, GGNN \cite{li2015gated}, DCNN \cite{atwood2016diffusion}, ChebyNet \cite{defferrard2016convolutional}, GCN \cite{kipf2016semi}, MPNN \cite{gilmer2017neural}, GraphSAGE \cite{hamilton2017inductive}, GPNN \cite{liao2018graph}, and GAT \cite{velivckovic2017graph}. Our proposed framework demonstrates a significant improvement of $25.35\%$ compared to the next-best baseline.

\vspace{-2mm}
\begin{table*}[h!]
\footnotesize
\centering
\def\arraystretch{1.05}
{%
\begin{tabular}{@{}c|cc@{}}
\toprule
Methods & Validation MAE ($\times 1.0e^{-3}$) & Test MAE ($\times 1.0e^{-3}$) \\
\midrule
GCN-FP ~\cite{duvenaud2015convolutional} & 15.06 $\pm$ 0.04 & 14.80 $\pm$ 0.09 \\
GGNN ~\cite{li2015gated} & 12.94 $\pm$ 0.05 & 12.67 $\pm$ 0.22 \\
DCNN ~\cite{atwood2016diffusion} & 10.14 $\pm$ 0.05 & 9.97 $\pm$ 0.09 \\
ChebyNet ~\cite{defferrard2016convolutional} & 10.24 $\pm$ 0.06 & 10.07 $\pm$ 0.09 \\
GCN ~\cite{kipf2016semi} & 11.68 $\pm$ 0.09 & 11.41 $\pm$ 0.10 \\
MPNN ~\cite{gilmer2017neural} & 11.16 $\pm$ 0.13 & 11.08 $\pm$ 0.11 \\
GraphSAGE ~\cite{hamilton2017inductive} & 13.19 $\pm$ 0.04 & 12.95 $\pm$ 0.11 \\
GPNN ~\cite{liao2018graph} & 12.81 $\pm$ 0.80 & 12.39 $\pm$ 0.77 \\
GAT ~\cite{velivckovic2017graph} & 11.39 $\pm$ 0.09 & 11.02 $\pm$ 0.06 \\
LanczosNet, ~\cite{liao2019lanczosnet} & 9.65 $\pm$ 0.19 & 9.58 $\pm$ 0.14 \\
AdaLanczosNet~\cite{liao2019lanczosnet} & 10.10 $\pm$ 0.22 & 9.97 $\pm$ 0.20 \\
\textbf{MMF W/GPT-4}  & \textbf{7.63 $\pm$ 0.07} & \textbf{7.45$\pm$ 0.03} \\
\bottomrule
\end{tabular}
}
\vspace{-1mm}
\caption{The table shows the experimental results of the framework's performance on the QM8 dataset in comparison to the baseline algorithms in terms of the MAE metric. We utilized the \texttt{MMF W/GPT-4} framework with the scaffold technique, setting $K$ to 16.}
\label{table:qm8}
\end{table*}

\vspace{-3mm}
Table \ref{tb:main_2} compares the framework's performance to baseline algorithms on the QM9 dataset.  The results are reported for the test dataset, using the mean absolute error (MAE) as the evaluation metric, with lower values indicating better performance. The baseline results are taken from a previous work\cite{gasteiger2020directional}. The baseline algorithms include SchNet (\cite{schutt2017schnet}), PhysNet (\cite{unke2019physnet}), Provably Powerful Graph Networks (PPGN, \cite{maron2019provably}), MEGNet-simple (\cite{chen2019graph}), Cormorant (C-Net, \cite{anderson2019cormorant}), and DimeNet (\cite{gasteiger2020directional}). Our framework demonstrates a significant improvement compared to the next-best baseline method.

\vspace{-2mm}
\begin{table}[h]
\footnotesize
\begin{adjustbox}{width=1.225\columnwidth,center}
\begin{tabular}{@{}c|c|ccccccc@{}}
\toprule
\addlinespace[0.05cm]
Target &                                             Unit &         PPGN\cite{maron2019provably} &                SchNet\cite{schutt2017schnet} &       PhysNet\cite{unke2019physnet} &     MEGNet-s\cite{chen2019graph} &    Cormorant\cite{anderson2019cormorant} &       DimeNet\cite{gasteiger2020directional} & \textbf{MMF W/GPT-4}\\
\hline
\addlinespace[0.05cm]
$\mu$                        &                                \si{D}       &  \num{0.047} &           \num{0.033} &  \num{0.0529} &   \num{0.05} &   \num{0.13} &   \text{\num{0.0286}} & \textbf{\num{0.0106}}\\
$\alpha$                     &                               \si{\bohr^3}       &  \num{0.131} &           \num{0.235} &  \num{0.0615} &  \num{0.081} &  \num{0.092} &  \text{\num{0.0469}}  & \textbf{\num{0.0219}} \\
$\epsilon_\text{HOMO}$       &                         \si{\milli\electronvolt} &   \num{40.3} &              \num{41} &    \num{32.9} &     \num{43} &     \num{36} &    \text{\num{27.8}} & \textbf{\num{18.43}}\\
$\epsilon_\text{LUMO}$       &                         \si{\milli\electronvolt} &   \num{32.7} &              \num{34} &    \num{24.7} &     \num{44} &     \num{36} &    \text{\num{19.7}}  &
\textbf{\num{9.57}}\\
$\Delta\epsilon$             &                         \si{\milli\electronvolt} &     \num{60.0} &              \num{63} &    \num{42.5} &     \num{66} &     \num{60} &    \textbf{\num{34.8}} & \textbf{\num{22.34}}\\
$\left< R^2 \right>$         &                                     \si{\bohr^2} &  \num{0.592} &  \textbf{\num{0.073}} &   \num{0.765} &  \num{0.302} &  \num{0.673} &            \num{0.331}  &\textbf{\num{0.108}}\\
ZPVE                         &                         \si{\milli\electronvolt} &   \num{3.12} &             \num{1.7} &    \num{1.39} &   \num{1.43} &   \num{1.98} &    \textbf{\num{1.29}} & \textbf{\num{0.785}}\\
$U_0$                        &                         \si{\milli\electronvolt} &   \num{36.8} &              \num{14} &    \num{8.15} &     \num{12} &     \num{28} &    \textbf{\num{8.02}} & \textbf{\num{3.55}}\\
$U$                          &                         \si{\milli\electronvolt} &   \num{36.8} &              \num{19} &    \num{8.34} &     \num{13} &            - &    \textbf{\num{7.89}} &\textbf{\num{2.43}} \\
$H$                          &                         \si{\milli\electronvolt} &   \num{36.3} &              \num{14} &    \num{8.42} &     \num{12} &            - &    \textbf{\num{8.11}} &  \textbf{\num{3.09}}\\
$G$                          &                         \si{\milli\electronvolt} &   \num{36.4} &              \num{14} &    \num{9.40} &     \num{12} &            - &    \textbf{\num{8.98}} & \textbf{\num{4.23}} \\
$c_\text{v}$ \vspace{1pt}    &  \si[per-mode=fraction]{\cal\per\mol\per\kelvin} &  \num{0.055} &           \num{0.033} &  \num{0.0280} &  \num{0.029} &  \num{0.031} &  \textbf{\num{0.0249}} & \textbf{\num{0.0137}}\\
\bottomrule
\end{tabular}
\end{adjustbox}
\vspace{1mm}
\caption{The table compares our method (on the right) to baselines (on the left) using MAE metric on the QM9 dataset. We utilized \texttt{MMF W/GPT-4} framework with scaffold technique, setting $K$=16.}
\label{tb:main_2}
\end{table}

\vspace{-9mm}
\section{Conclusion}
\vspace{-4mm}
In the rapidly evolving field of computational chemistry, the pressing need for methodologies with higher accuracy and robustness in predicting molecular properties is undeniable. Our pioneering efforts in this study have introduced the Multi-Modal Fusion (MMF) framework, which synergistically amalgamates LLMs and GNNs to enhance the accuracy of molecular property predictions. Our approach not only improves predictions but also reduces the likelihood of overfitting, surpassing existing benchmarks in property prediction tasks. Our results on benchmark datasets confirm our hypothesis that fusing information from text and graph-based modalities can significantly enhance performance. This breakthrough opens new avenues for scientific discovery, advancing computational chemistry applications across domains to shape the next generation of tools and insights in chemistry.

\newpage

\section{Technical appendix}

\vspace{-3mm}
\subsection{Extended Experimental Results}
\vspace{-2mm}
Table \ref{tb:main_3} and \ref{tb:main_4} show the performance of our \texttt{MMF} framework when paired with different off-the-shelf LLMs on the QM8 and QM9 datasets, respectively. The underlying hypothesis of our framework is that GNNs can be utilized for initial explorations and generating baseline results in this regard. Zero-Shot CoT prompting of LLMs can be harnessed to enhance the preliminary outcomes of GNNs by utilizing the implicit domain-specific knowledge embedded within LLMs trainable parameters to obtain expressive cross-modal embeddings. Few-Shot ICL, on the other hand, can be utilized to further refine the framework's predictions by providing demonstrations from the training data, potentially leading to a more robust and accurate predictive framework for molecular property prediction. The experimental findings support the validity of this hypothesis, advancing drug discovery and materials science --- a task where conventional deep learning methods often fall short. This glaring gap in the integration of graph-based and linguistic insights renders current architectures less comprehensive and nuanced, potentially impeding breakthroughs across various disciplines.

\vspace{-2mm}
\begin{table*}[h!]
\footnotesize
\centering
\def\arraystretch{0.95}
{%
\begin{tabular}{@{}c|cc@{}}
\toprule
Methods & Validation MAE ($\times 1.0e^{-3}$) & Test MAE ($\times 1.0e^{-3}$) \\
\midrule
\text{MMF W/GPT-4}  & \textbf{7.63 $\pm$ 0.07} & \textbf{7.45$\pm$ 0.03} \\
\text{MMF W/GPT-3.5-turbo}  & \text{8.13 $\pm$ 0.09} & \text{8.21 $\pm$ 0.03} \\
\text{MMF W/GPT-3.0-text-davinci-003}  & \text{8.28 $\pm$ 0.06} & \text{8.37 $\pm$ 0.07} \\
\text{MMF W/Google Bard} & \text{9.15$\pm$ 0.04} & \text{9.33 $\pm$ 0.06} \\
\bottomrule
\end{tabular}
}
\vspace{-1mm}
\caption{The table shows the  \texttt{MMF} framework performance with different off-the-shelf LLMs on the QM8 dataset using the MAE metric. All the experiments were performed with the scaffold technique, setting $K$ to 16.}
\label{tb:main_3}
\vspace{-2mm}
\end{table*}

\vspace{-2mm}
\begin{table}[h]
\footnotesize
\begin{adjustbox}{width=1.225\columnwidth,center}
\begin{tabular}{@{}c|c|ccccc@{}}
\toprule
\addlinespace[0.05cm]
Target &                                             Unit &         \textbf{MMF W/GPT-4} &               \textbf{MMF W/GPT-3.5-turbo} &      \textbf{MMF W/GPT-3.0-text-davinci-003} &     \textbf{MMF W/Google Bard} &    \\
\hline
\addlinespace[0.05cm]
$\mu$                        &                                \si{D}       &  \num{0.0106} &           \num{0.0124} &  \num{0.0127} &   \num{0.0189}\\
$\alpha$                     &                               \si{\bohr^3}       &  \num{0.0219} &           \num{0.0237} &  \num{0.0249} & \num{0.0354}  \\
$\epsilon_\text{HOMO}$       &                         \si{\milli\electronvolt} &   \num{18.43} &           \num{20.43} &  \num{21.47} &  \num{24.97} \\
$\epsilon_\text{LUMO}$       &                         \si{\milli\electronvolt} &   \num{9.57} &           \num{10.02} &  \num{11.85} &  \num{15.65} \\
$\Delta\epsilon$             &                         \si{\milli\electronvolt} &    \num{22.34} &           \num{24.08} &  \num{25.97} &  \num{29.04} \\
$\left< R^2 \right>$         &                                     \si{\bohr^2} &  \num{0.108} &           \num{0.137} &  \num{0.142} &  \num{0.287} \\
ZPVE                         &                         \si{\milli\electronvolt} &   \num{0.0785} &           \num{0.0846} &  \num{0.0905} & \num{1.143}  \\
$U_0$                        &                         \si{\milli\electronvolt} &   \num{3.55} &           \num{4.19} &  \num{4.52} &  \num{6.74}  \\
$U$                          &                         \si{\milli\electronvolt} &  \num{2.43} &           \num{3.16} &  \num{3.97} &  \num{5.89} \\
$H$                          &                         \si{\milli\electronvolt} &   \num{3.09} &           \num{3.48} &  \num{3.72} &  \num{5.63} \\
$G$                          &                         \si{\milli\electronvolt} &   \num{4.23} &           \num{4.71} &  \num{4.89} &  \num{6.67} \\
$c_\text{v}$ \vspace{1pt}    &  \si[per-mode=fraction]{\cal\per\mol\per\kelvin} &  \num{0.0137} &           \num{0.0144} &  \num{0.0158} &   \num{0.0217}\\
\bottomrule
\end{tabular}
\end{adjustbox}
\vspace{1mm}
\caption{The table shows the  \texttt{MMF} framework performance using various off-the-shelf LLMs on QM9 dataset in terms of MAE metric. All experiments utilized the scaffold technique, with 
$K$ set to 16.}
\label{tb:main_4}
\end{table}

\vspace{-8mm}
\subsection{Additional Datasets and Experimental Results}
\vspace{-2mm}
We have additionally evaluated the performance of our multi-modal learning framework on four benchmark datasets: ESOL\cite{delaney2004esol}, FreeSolv\cite{mobley2014freesolv}, Lipophilicity\cite{gaulton2017chembl}, and PDBbind\cite{wang2004pdbbind}, comparing it against several popular algorithms. Each of these datasets plays a critical role in various areas of computational chemistry and molecular modeling, particularly in the field of drug discovery and development. We partitioned the data into multiple sets of 80$\%$ for training, 10$\%$ for validation, and 10$\%$ for testing. To ensure that molecules with the same scaffold did not appear in multiple splits, we utilized scaffold splitting, following the strategy implemented in \cite{yuksel2023selformer} using the Chemprop library\cite{yang2019analyzing}. This approach offers a more resilient and robust evaluation of the framework's performance compared to random splitting. Table \ref{tb:mainad1} presents the experimental results of the framework performance compared to the baselines. Our experimental findings support the efficacy of the $\texttt{MMF}$ framework, which outperformed other state-of-the-art baselines on all four datasets, achieving lower prediction error scores. The baseline results are reported from a previous study\cite{yuksel2023selformer}. Moreover, our framework can identify subtle differences in molecular structures, enhancing the expressiveness of cross-modal embeddings for improved property prediction. We aim to further explore the potential of our multi-modal learning framework in classification-based molecular property prediction tasks. Thus, we evaluated our $\texttt{MMF}$ framework performance in comparison to the baselines using a set of additional benchmark datasets, including {BBBP}, {HIV}, {BACE}, {Tox21}, and {ClinTox}, which were originally introduced in an earlier study\cite{wu2018moleculenet}. These datasets comprise chemical SMILES representations paired with binary labels that indicate specific molecular properties, such as toxicity or the ability to inhibit HIV replication. The baseline results have been reported from a previous study\cite{yuksel2023selformer}. Performance is measured using the ROC curve metric, with higher scores indicating better results. Entries marked with ``-" denote unavailable data for specific method-task combinations. The BBBP and BACE benchmark datasets are balanced, whereas the other datasets are highly imbalanced. For in-context learning (ICL), we select samples either randomly or based on the highest Tanimoto similarity computed using RDKit; however, the latter method does not guarantee a balanced class distribution. Our study employs a flexible sampling strategy, aiming to obtain a representative sample from datasets with high label imbalances. Specifically, our strategy assists in sampling demonstrations for constructing augmented prompts, utilizing a 3:2 majority-to-minority class ratio for framework evaluations. Table \ref{tb:mainad2} showcases the performance of the framework in comparison to the baselines.

\vspace{-5mm}
\begin{table}[h]
\small
\caption{Our proposed framework was evaluated against various baselines for molecular property predictions using several datasets, such as ESOL, FreeSolv, Lipophilicity, and PDBbind. The performance was measured using RMSE, with lower values indicating better prediction accuracy. All the experiments were performed with the scaffold technique, setting $K$ to 16.}
\label{tb:mainad1}
\centering
\vspace{0mm}
{\begin{tabular}{@{}lllll@{} }
\toprule
\addlinespace[0.05cm]
& \textbf{ESOL} & \textbf{FreeSolv} & \textbf{Lipophilicity} & \textbf{pdbbind}  \\
\addlinespace[0.05cm]
\hline  
\addlinespace[0.05cm]
\textbf{MMF W/GPT-4} & \textbf{0.413} & \textbf{1.283} & \textbf{0.373} & \textbf{1.028} \\
\textbf{MMF W/GPT-3.5-turbo} & \text{0.457} & \text{1.374} & \text{0.427} & \text{1.185} \\
\textbf{MMF W/GPT-3.0-text-davinci-003} & \text{0.513} & \text{1.408} & \text{0.459} & \text{1.213} \\
\textbf{MMF W/Google Bard} & \text{0.608} & \text{1.675} & \text{0.553} & \text{1.343}\\
\hline
\addlinespace[0.05cm]
\addlinespace[0.05cm]
SELFormer\cite{yuksel2023selformer} & 0.682 &
2.797 &
0.735 &
1.488 \\
\addlinespace[0.05cm]
D-MPNN\cite{yang2019analyzing} & 1.050 &
2.082 &
0.683 &
1.397 \\
\addlinespace[0.01cm]
MolCLR\cite{fang2022geometry} & 1.110 &
2.200 &
0.650 &
- \\
\addlinespace[0.01cm]
Hu et al.\cite{hu2019strategies} & 1.220 &
2.830 &
 0.740 &
- \\
\addlinespace[0.01cm]
MGCN\cite{lu2019molecular} & 1.270 &
3.350 &
1.110 &
- \\
\addlinespace[0.01cm]
GEM\cite{fang2022geometry} & 0.798 &
1.877 &
0.660 &
- \\
\addlinespace[0.01cm]
SchNet\cite{schutt2017schnet} & 1.050 &
3.220 &
0.910 &
- \\
\addlinespace[0.01cm]
KPGT\cite{li2022kpgt} & 0.803 &
2.121 &
0.600 &
- \\
\addlinespace[0.01cm]
GraphMVP-C\cite{liu2021pre} & 1.029 &
- &
 0.681 &
- \\
\addlinespace[0.01cm]
GCN\cite{kipf2016semi} & 1.430 &
2.870 &
 0.850 &
- \\
\addlinespace[0.01cm]
GIN\cite{xu2018powerful} & 1.450 &
2.760 & 
0.850 &
- \\
\addlinespace[0.01cm]
ChemBERTa-2\cite{ahmad2022chemberta}  & - &
- &
0.986 &
- \\
\addlinespace[0.1cm]
\bottomrule
\end{tabular}}{}
\vspace{-4mm}
\end{table}

\vspace{-3mm}
\begin{table}[h]
\small
\caption{We compared our proposed framework with various baselines for predicting molecular properties using datasets, including BACE, BBBP, HIV, Tox21, and SIDER. The performance was evaluated using the ROC curve metric, where higher scores indicate better results. All the experiments were performed with the scaffold technique, setting $K$ to 16.}
\label{tb:mainad2}
\centering
\vspace{0mm}
{\begin{tabular}{@{}llllll@{} }
\toprule
\addlinespace[0.1cm]
& \textbf{BACE} & \textbf{BBBP} & \textbf{HIV} & \textbf{Tox21} & \textbf{SIDER} \\
\addlinespace[0.1cm]
\hline
\addlinespace[0.1cm]
 & ROC & ROC & ROC & ROC & ROC  \\
 \addlinespace[0.1cm]
 \addlinespace[0.05cm]
\hline  
\addlinespace[0.05cm]
\textbf{MMF W/GPT-4} & \textbf{0.893} & \textbf{0.937} & \textbf{0.862} & \textbf{0.898}  & \textbf{0.812} \\
\textbf{MMF W/GPT-3.5-turbo} & \text{0.881} & \text{0.929} & \text{0.851} & \text{0.877} & \text{0.809} \\
\textbf{MMF W/GPT-3.0-text-davinci-003} & \text{0.877} & \text{0.921} & \text{0.849} & \text{0.872} & \text{0.792} \\
\textbf{MMF W/Google Bard} & \text{0.861} & \text{0.916} & \text{0.826} & \text{0.859} & \text{0.766}\\
\hline
\addlinespace[0.05cm]
SELFormer\cite{yuksel2023selformer} & 0.832 &
\text{0.902} &
0.681 &
0.653 &
\text{0.745} \\
D-MPNN\cite{yang2019analyzing}  & 0.809 & 0.710 & 0.771 & 0.759 & 0.570
\\
\addlinespace[0.01cm]
MolBERT\cite{fabian2020molecular} & 0.866 & 0.762 & 0.783 & - & - 
\\
\addlinespace[0.01cm]
ChemBERTa-2\cite{ahmad2022chemberta}  & 0.799 & 0.728 & 0.622 & - & -
\\
\addlinespace[0.01cm]
Hu et al.\cite{hu2019strategies} & 0.859 & 0.708 & 0.802 & 0.787 & 0.652
\\
\addlinespace[0.01cm]
MolCLR\cite{fang2022geometry} &  \text{0.890} & 0.736 & \text{0.806} & 0.787 & 0.652
\\
\addlinespace[0.01cm]
GraphMVP-C\cite{liu2021pre} & 0.812 & 0.724 & 0.770 & 0.744 & 0.639
\\
\addlinespace[0.01cm]
GEM\cite{fang2022geometry} & 0.856 & 0.724 & \text{0.806} & 0.781 & 0.672
\\
\addlinespace[0.01cm]
MGCN\cite{lu2019molecular} & 0.734 & 0.850 & 0.738 & 0.707 & 0.552
\\
\addlinespace[0.01cm]
GCN\cite{kipf2016semi} & 0.716 & 0.718 & 0.740 & 0.709 & 0.536 
\\
\addlinespace[0.01cm]
GIN\cite{xu2018powerful} & 0.701 & 0.658 & 0.753 & 0.740 & 0.573 
\\
\addlinespace[0.01cm]
SchNet\cite{schutt2017schnet} & 0.766 & 0.848 & 0.702 & 0.772 & 0.539
\\
\addlinespace[0.01cm]
KPGT\cite{li2022kpgt} & 0.855 & \text{0.908} & - & \text{0.848} & 0.649
\\
\addlinespace[0.1cm]
\bottomrule
\end{tabular}}{}
\vspace{-1mm}
\end{table}

\vspace{-4mm}
\subsection{Ablation Study}
\vspace{-2mm}
Our proposed framework obtains unified embeddings by integrating knowledge from Zero-Shot

\newpage
CoT and Few-Shot ICL learning methods, utilizing prompting based on LLMs (refer to Subsection \ref{lms}) and GNNs (refer to Subsection \ref{gcc}). We conducted ablation studies to examine the impact of each method on the overall enhanced performance of our framework. By selectively disabling methods, we created multiple ablated variants of our framework and evaluated them using benchmark datasets for the property prediction task. This approach enabled us determine the contributions of the disabled methods to the overall framework performance. We chose the proposed $\texttt{MMF}$ framework as the reference baseline for the ablation studies. This rigorous approach not only validates the effectiveness of the different methods but also provides justification for their design choices and inclusion within the framework. Our proposed framework operates through a multi-step pipeline, as follows:

\vspace{-3mm}
 \begin{itemize}
\item[\ding{51}] \text{Synergistic Cross-Modal Embedding Generation (SEG)}: The Zero-Shot CoT prompts LLMs to generate technical descriptions of chemical SMILES representations. These descriptions are then used to fine-tune smaller LMs for domain-specific customization, facilitating the computation of context-aware token embeddings. We utilize softmax attention pooling to obtain text-level embeddings from these contextualized token embeddings. Simultaneously and in parallel, the Graph Chebyshev Convolution operator computes graph-level embeddings. These two sets of embeddings are synergistically integrated using a multi-head attention mechanism, thereby establishing a robust framework for integrating structured and unstructured data.
\vspace{-1mm}
\item[\ding{51}] \text{Predictive Embedding Generation (PEG)}: The Few-Shot ICL method prompts LLMs to predict molecular properties based on a few demonstrations of the downstream task, which are then subsequently encoded to obtain prediction embeddings. ICL guides predictions without requiring explicit fine-tuning, relying solely on the implicit pre-trained knowledge encapsulated within the LLMs' parameters, and conditioned on the context-augmented prompts.
\vspace{-1mm}
\item[\ding{51}] \text{MOE Dynamic Prediction (MOE-DP)}: For the output layer, we employ the MOE technique with a gating mechanism, where cross-modal and prediction embeddings are unified through the gating mechanism. The goal is to optimize predictive performance and fine-tune the embeddings, leveraging diverse learning strategies for precise molecular property predictions.
\end{itemize}

\vspace{-2mm}
The ablated variants without the synergistic embedding generation (SEG), predictive embedding generation (PEG), and MOE dynamic prediction (MOE-DP) methods are referred to as `w/o SEG,' `w/o PEG,' and `w/o MOE-DP,' respectively. In the case of `w/o MOE-DP,' we utilize a linear operator to predict the molecular properties. The experimental findings from the ablation study are shown in Tables \ref{tb:main_5} and \ref{tb:main_6}. All experiments were conducted using the scaffold technique with $K$=16. In the ablation study focusing on molecular property estimation, we evaluated the performance of the ablated variants against the baseline using the MAE error metric, allowing for a comprehensive analysis. Upon examination, it becomes evident that the synergistic embedding generation (SEG) method within the \texttt{MMF} framework holds greater significance than the predictive embedding generation (PEG) and MOE dynamic prediction (MOE-DP) methods for attaining state-of-the-art performance on the benchmark datasets. For the QM8 dataset, the `w/o SEG' variant shows a substantial decline in performance relative to the baseline, as evidenced by a marked increase of $46.44\%$ in MAE. In contrast, the `w/o PEG' variant exhibits marginally inferior performance compared to the baseline, with a modest increase of $14.49\%$ in MAE. Similarly, the `w/o MOE-DP' variant performs much worse than the baseline, with an increase of $20.41\%$ in MAE. This increase in error might be attributed to the substitution with an oversimplified linear operator in the output layer. Similar trends were observed in the QM9 dataset. The higher increase in the performance metrics of the ablated variants, when compared to the baseline, underscores the relative significance of the mechanisms underpinning the omitted methods of the baseline. The experimental findings suggest that integrating knowledge from both text and graph modalities is a promising approach for enhancing the overall performance of the framework. Specifically, our innovative framework leverages a fusion of deep learning techniques, including GNNs and both larger and smaller language models, to predict molecular properties. The zero-shot CoT prompting of LLMs generates technical descriptions of molecules, which are then encoded by smaller language models to generate text-level embeddings. These text-level embeddings are seamlessly integrated with graph embeddings, resulting in improved cross-modal embeddings. We jointly optimize the cross-modal embeddings and the prediction embeddings generated by the few-shot ICL prompting of LLMs through a MOE technique with gating mechanism and then combine them to enhance the performance of our framework. In conclusion, this holistic approach not only demonstrates the benefits of combining diverse knowledge sources for optimized molecular property predictions but also provides a comprehensive understanding of molecular representations, setting a benchmark in the domain.

\vspace{-2mm}
\begin{table*}[h!]
\footnotesize
\begin{adjustbox}{width=0.65\columnwidth,center}
\centering
{%
\begin{tabular}{@{}c|cc@{}}
\toprule
\addlinespace[0.05cm]
Methods & Validation MAE ($\times 1.0e^{-3}$) & Test MAE ($\times 1.0e^{-3}$) \\
\midrule
\textbf{MMF W/GPT-4}  & \textbf{7.63 $\pm$ 0.07} & \textbf{7.45$\pm$ 0.03} \\
w/o SEG & 10.86 $\pm$ 0.02 & 10.91 $\pm$ 0.05 \\
w/o PEG &  8.47$\pm$ 0.03 &  8.53$\pm$ 0.02 \\
w/o MOE-DP & 8.89$\pm$ 0.04 & 8.97 $\pm$ 0.03 \\
\bottomrule
\end{tabular}
}
\end{adjustbox}
\vspace{-2mm}
\caption{The table presents the ablation study results on the QM8 dataset.}
\label{tb:main_5}
\vspace{-2mm}
\end{table*}

\vspace{-2mm}
\begin{table}[h]
\footnotesize
\begin{adjustbox}{width=0.8\columnwidth,center}
\begin{tabular}{@{}c|c|ccccc@{}}
\toprule
\addlinespace[0.075cm]
Target &                                             Unit &         \textbf{MMF W/GPT-4}  &               w/o SEG &      w/o PEG &     w/o MOE-DP & \\
\hline
\addlinespace[0.05cm]
$\mu$                        &                                \si{D}       &  \num{0.0106} &           \num{0.0251} &  \num{0.0124} &  \num{0.0179} \\
$\alpha$                     &                               \si{\bohr^3}       &  \num{0.0219} &           \num{0.0456} &  \num{0.0298} & \num{0.0365} \\
$\epsilon_\text{HOMO}$       &                         \si{\milli\electronvolt} &   \num{18.43} &           \num{29.36} &  \num{21.05} &  \num{25.98} \\
$\epsilon_\text{LUMO}$       &                         \si{\milli\electronvolt} &   \num{9.57} &           \num{19.91} &  \num{12.43} &  \num{15.77}\\
$\Delta\epsilon$             &                         \si{\milli\electronvolt} &    \num{22.34} &           \num{32.91} &  \num{24.57} &  \num{27.09}\\
$\left< R^2 \right>$         &                                     \si{\bohr^2} &  \num{0.108} &           \num{0.278} &  \num{0.143} &   \num{0.221}\\
ZPVE                         &                         \si{\milli\electronvolt} &   \num{0.0785} &           \num{1.35} &  \num{0.096} & \num{1.14}  \\
$U_0$                        &                         \si{\milli\electronvolt} &   \num{3.55} &           \num{7.81} &  \num{4.48} &  \num{6.29} \\
$U$                          &                         \si{\milli\electronvolt} &  \num{2.43} &           \num{7.76} &  \num{3.14} & \num{5.97}  \\
$H$                          &                         \si{\milli\electronvolt} &   \num{3.09} &           \num{8.24} &  \num{4.15} & \num{5.04} \\
$G$                          &                         \si{\milli\electronvolt} &   \num{4.23} &           \num{9.32} &  \num{5.17} &  \num{6.83}\\
$c_\text{v}$ \vspace{1pt}    &  \si[per-mode=fraction]{\cal\per\mol\per\kelvin} &  \num{0.0137} &           \num{0.0311} &  \num{0.0162} &  \num{0.0253} \\
\bottomrule
\end{tabular}
\end{adjustbox}
\vspace{1mm}
\caption{The table presents the ablation study results on the QM9 dataset.}
\label{tb:main_6}
\vspace{-4mm}
\end{table}

\vspace{-4mm}
\subsection{Additional In-Depth Studies}
\vspace{-3mm}
In this section, we will analyze more into the synergistic cross-modal embedding generation(SEG) and predictive embedding generation(PEG) methods. 

\vspace{-3mm}
\subsubsection{Impact of Synergistic Cross-Modal Embedding Generation(SEG) method}
\vspace{-2mm}
The synergistic embedding generation(SEG) method consists of three main components:

\vspace{-2mm}
\begin{itemize}
\item[\ding{51}] Text-Level Embeddings (TL-Emb): Computed using softmax attention pooling of contextual token embeddings. These context-aware embeddings are obtained from fine-tuning smaller LMs on the technical descriptions generated by the zero-shot CoT prompting of LLMs on chemical SMILES representations.
\item[\ding{51}] Graph-Level Embeddings (GL-Emb): Computed through Graph Chebyshev Convolution technique on the corresponding non-linear molecular graph representations obtained from the chemical SMILES representations.
\item[\ding{51}] Cross-Modal Embeddings (CM-Emb): Computed using a multi-head attention mechanism to facilitate the robust integration and analysis of text-level and graph-level embeddings.
\end{itemize}

\vspace{-2mm}
We conducted ablation studies to examine the impact of different embeddings within the Synergistic Embedding Generation (SEG) method on the overall improved performance of our framework. The ablated variants without the text-level embeddings (TL-Emb), graph-level embeddings (GL-Emb), and cross-modal embeddings (CM-Emb) methods are referred to as `w/o TL-Emb', `w/o GL-Emb', and `w/o CM-Emb', respectively. In the case of `w/o CM-Emb', we concatenate the embeddings and utilize a linear operator to predict the cross-modal embeddings.
The experimental findings from the ablation study are presented in Tables \ref{tb:main_7} and \ref{tb:main_8}. 

\vspace{-2mm}
\begin{table*}[h!]
\footnotesize
\begin{adjustbox}{width=0.675\columnwidth,center}
\centering
{%
\begin{tabular}{@{}c|cc@{}}
\toprule
Methods & Validation MAE ($\times 1.0e^{-3}$) & Test MAE ($\times 1.0e^{-3}$) \\
\midrule
\textbf{MMF W/GPT-4}  & \textbf{7.63 $\pm$ 0.07} & \textbf{7.45$\pm$ 0.03} \\
w/o TL-Emb &  9.43$\pm$ 0.06 &  9.51$\pm$ 0.02 \\
w/o GL-Emb &  8.11$\pm$ 0.04 &  8.16$\pm$ 0.07 \\
w/o CM-Emb &  10.15$\pm$ 0.03 &  10.27$\pm$ 0.05 \\
\bottomrule
\end{tabular}
}
\end{adjustbox}
\vspace{-1mm}
\caption{The table presents the impact of various embeddings within the synergistic embedding generation (SEG) method on the QM8 dataset. All experiments were conducted using the scaffold technique with $K$=16.}
\label{tb:main_7}
\end{table*}

\vspace{-3mm}
Upon closer examination, it becomes evident that text-level embeddings (TL-Emb) are more significant than graph-level embeddings (GL-Emb) in achieving state-of-the-art performance on the benchmark datasets. For the QM8 dataset, the `w/o TL-Emb' variant demonstrates a substantial decline in performance compared to the baseline, as evidenced by a significant increase of $27.65\%$ in MAE. In contrast, the `w/o GL-Emb' variant exhibits only a marginally inferior performance compared to the baseline, with a modest increase of $9.53\%$  in MAE. The `w/o CM-Emb' variant shows the poorest performance relative to the baseline with an increase of $37.85\%$ in MAE. This increase in error may be attributed to the substitution of an oversimplified concatenation of text-level and graph-level embeddings, along with the utilization of a linear operator to predict the cross-modal embeddings. Similar trends were observed in the QM9 dataset. The greater increase in performance metrics for the ablated variants compared to the baseline underscores the relative significance of the omitted methods

\vspace{-2mm}
\begin{table}[h]
\footnotesize
\begin{adjustbox}{width=0.8\columnwidth,center}
\begin{tabular}{@{}c|c|ccccc@{}}
\toprule
\addlinespace[0.05cm]
Target &                                             Unit &         \textbf{MMF W/GPT-4}  &               w/o TL-Emb &      w/o GL-Emb &     w/o CM-Emb & \\
\hline
\addlinespace[0.05cm]
$\mu$                        &                                \si{D}       &  \num{0.0106} &           \num{0.0158} &  \num{0.139} &  \num{0.257} \\
$\alpha$                     &                               \si{\bohr^3}       &  \num{0.0219} &           \num{0.0337} &  \num{0.0308} & \num{0.0413} \\
$\epsilon_\text{HOMO}$       &                         \si{\milli\electronvolt} &   \num{18.43} &           \num{24.19} &  \num{23.31} &   \num{26.97}\\

$\epsilon_\text{LUMO}$       &                         \si{\milli\electronvolt} &   \num{9.57} &           \num{13.96} &  \num{12.28} &  \num{18.96}\\
$\Delta\epsilon$             &                         \si{\milli\electronvolt} &    \num{22.34} &           \num{27.65} &  \num{26.89} &  \num{31.07}\\
$\left< R^2 \right>$         &                                     \si{\bohr^2} &  \num{0.108} &           \num{0.146} &  \num{0.153} &   \num{0.292}\\

ZPVE                         &                         \si{\milli\electronvolt} &   \num{0.0785} &           \num{0.098} &  \num{0.087} &  \num{1.21} \\
$U_0$                        &                         \si{\milli\electronvolt} &   \num{3.55} &           \num{5.17} &  \num{4.92} &   \num{7.88}\\
$U$                          &                         \si{\milli\electronvolt} &  \num{2.43} &           \num{3.19} &  \num{3.37} &  \num{7.71} \\

$H$                          &                         \si{\milli\electronvolt} &   \num{3.09} &           \num{4.16} &  \num{3.97} & \num{7.75} \\
$G$                          &                         \si{\milli\electronvolt} &   \num{4.23} &           \num{5.05} &  \num{4.92} & \num{7.89} \\
$c_\text{v}$ \vspace{1pt}    &  \si[per-mode=fraction]{\cal\per\mol\per\kelvin} &  \num{0.0137} &           \num{0.0187} &  \num{0.0172} &   \num{0.0223}\\
\bottomrule
\end{tabular}
\end{adjustbox}
\vspace{1mm}
\caption{The table showcases the influence of diverse embeddings in the synergistic embedding generation (SEG) method on the QM9 dataset. All experiments were conducted using the scaffold technique with $K$=16.}
\vspace{-6mm}
\label{tb:main_8}
\end{table}

\vspace{-3mm}
\subsubsection{Impact of Predictive Embedding Generation(PEG) method}
\vspace{-2mm}
In-context learning (ICL) or few-shot prompting enables pretrained foundational large language models (LLMs) to adapt to new tasks with only a few task-specific demonstrations, thus eliminating the need for parameter updates. This approach represents a departure from traditional supervised learning methods. In our study, we employ few-shot prompting with LLMs to predict molecular properties of new, unseen molecules. We construct context-augmented prompts, comprising task-specific instructions and demonstrations (inputs in the form of chemical SMILES strings and outputs as molecular property pairs) sampled from the training data to direct general-purpose LLMs in predicting molecular properties of the target chemical SMILES strings. In essence, the instruction serves to contextualize the task of predicting molecular properties for the target chemical SMILES strings, while the demonstrations are crucial in guiding the language model to produce relevant and accurate responses for the specific task at hand. In this scenario, the augmented prompt guides the LLMs to tap into the pre-existing knowledge embedded within their parameters, acquired during training on vast and diverse corpora, to predict the molecular graph properties. We explore two sampling strategies, "Random" and "Scaffold", to construct augmented prompts, and we evaluate the predictive abilities of LLMs that rely solely on these prompts. Our experiments focus on both the quality (how helpful are the demonstrations in reducing prediction error) and quantity (number of demonstrations) of these sampling methods, aiming to enhance property prediction accuracy. 

\vspace{-3mm}
\paragraph{Results:} Tables \ref{tb:main_9} and \ref{tb:main_10} present the MAE scores, showcasing the results of property prediction performance in our study on the impact of both quality and quantity of demonstrations during few-shot prompting of LLMs. Our analysis revealed that the \texttt{MMF W/GPT} models outperformed the \texttt{MMF W/Google Bard} in terms of MAE on both datasets under examination. Notably, among the evaluated \texttt{MMF W/GPT} models, \texttt{MMF W/GPT-4} exhibited superior performance compared to \texttt{MMF W/davinci-003} and \texttt{MMF W/GPT-3.5 Turbo} in predicting molecular properties. Our study supports the notion that increasing training examples in few-shot prompting(in-context learning) can significantly enhance framework performance. It underscores a direct correlation between the volume of ICL demonstrations and the predictive accuracy of the LLMs. Furthermore, our investigation provides compelling evidence that $\text{scaffold sampling}$ consistently outperforms random sampling across distinct datasets (QM8 and QM9), further bolstering the empirical validity of our research.  A possible reason for this might be the structural similarities between the molecules sampled from the scaffold strategy and the query molecule. This could potentially tilt the \texttt{MMF GPT} models towards making more precise decisions. LLMs still face a significant limitation in understanding molecular representations in SMILES strings, resulting in inaccurate results in property prediction tasks. SMILES notation is a widely used textual representation for chemical structures, but LLMs struggle to interpret it accurately due to issues like implicit hydrogen atoms, multiple valid representations for a single molecule, which leads to ambiguity, and treating SMILES strings as mere sequences of characters. This hampers the performance of LLMs in tasks such as property prediction and affects downstream cheminformatics tasks. Therefore, LLMs with improved capabilities in handling molecular structures and coupling with existing tools such as RDKit will be necessary. 

\vspace{-3mm}
\begin{table*}[h!]
\small
\begin{adjustbox}{width=0.80\columnwidth,center}
\centering
\def\arraystretch{1.25}
{%
\begin{tabular}{@{}c|cc@{}}
\toprule
Methods & Validation MAE ($\times 1.0e^{-3}$) & Test MAE ($\times 1.0e^{-3}$) \\
\midrule
MMF W/ GPT-4 (Scaffold, $K$=4) &  \text{9.83 $\pm$ 0.05} & \text{9.89$\pm$ 0.07} \\
MMF W/ GPT-4 (Scaffold, $K$=12) &  \text{8.24 $\pm$ 0.04} & \text{8.36$\pm$ 0.02} \\
MMF W/ GPT-4 (random, $K$=12) &  \text{10.67 $\pm$ 0.03} & \text{10.72$\pm$ 0.06} \\
\hline
MMF W/ GPT-3.5 (Scaffold, $K$=4) &  \text{10.35 $\pm$ 0.04} & \text{10.41 $\pm$ 0.06} \\
MMF W/ GPT-3.5 (Scaffold, $K$=12) &  \text{8.93 $\pm$ 0.08} & \text{8.98 $\pm$ 0.05} \\
MMF W/ GPT-3.5 (random, $K$=12) &  \text{11.03 $\pm$ 0.07} & \text{11.17 $\pm$ 0.03} \\
\hline
MMF W/ davinci-003 (Scaffold, $K$=4)  &  \text{10.13 $\pm$ 0.03} & \text{10.09 $\pm$ 0.06} \\
MMF W/ davinci-003 (Scaffold, $K$=12)  &  \text{8.87 $\pm$ 0.07} & \text{8.96 $\pm$ 0.09} \\
MMF W/ davinci-003 (random, $K$=12)   &  \text{10.98 $\pm$ 0.05} & \text{11.02 $\pm$ 0.04} \\
\hline
MMF W/ Google Bard (Scaffold, $K$=12)  &  \text{9.53$\pm$ 0.04} & \text{9.67 $\pm$ 0.03} \\
MMF W/ Google Bard (random, $K$=12)   &  \text{11.45$\pm$ 0.05} & \text{11.63 $\pm$ 0.07} \\
\bottomrule
\end{tabular}
}
\end{adjustbox}
\vspace{-1mm}
\caption{The table shows the MAE scores of the \texttt{MMF W/GPT} models for predicting molecular properties on the QM8 dataset. Here, $K$ represents the number of training examples used in few-shot prompting. All experiments were conducted using the scaffold technique with $K$=16.}
\label{tb:main_9}
\vspace{-2mm}
\end{table*}

\vspace{-2mm}
\begin{table}[h]
\footnotesize
\begin{adjustbox}{width=1.4\columnwidth,center}
\begin{tabular}{@{}c|c|ccccc@{}}
\toprule
Target &                                             Unit &                     MMF W/ GPT-4 (Scaffold, $K$=16) &       MMF W/ GPT-4 (Scaffold, $K$=4)  &     MMF W/GPT-davinci (Scaffold, $K$=16) &     MMF W/GPT-davinci (Scaffold, $K$=4) \\
\midrule
$\mu$                        &                                \si{D}       &  \num{0.0106} &           \num{0.0132} &  \num{0.0127} &  \num{0.0137}\\
$\alpha$                     &                               \si{\bohr^3}       &  \num{0.0219} &          \num{0.0443} &  \num{0.0249} & \num{0.0483}  \\
$\epsilon_\text{HOMO}$       &                         \si{\milli\electronvolt} &   \num{18.43} &            \num{26.02} &  \num{21.47} &  \num{32.39}\\
$\epsilon_\text{LUMO}$       &                         \si{\milli\electronvolt} &   \num{9.57} &          \num{17.58} &  \num{11.85} &   \num{19.42} \\
$\Delta\epsilon$             &                         \si{\milli\electronvolt} &    \num{22.34} &           \num{30.43} &  \num{25.97} &  \num{32.51} \\
$\left< R^2 \right>$         &                                     \si{\bohr^2} &  \num{0.108} &           \num{0.216} &  \num{0.142} &  \num{0.253} \\
ZPVE                         &                         \si{\milli\electronvolt} &   \num{0.0785} &           \num{0.095} &  \num{0.0905} & \num{1.073}  \\
$U_0$                        &                         \si{\milli\electronvolt} &   \num{3.55} &            \num{5.34} &  \num{4.52} &  \num{6.05}  \\
$U$                          &                         \si{\milli\electronvolt} &  \num{2.43} &           \num{4.92} &  \num{3.97} &  \num{6.89} \\
$H$                          &                         \si{\milli\electronvolt} &   \num{3.09} &           \num{5.01} &  \num{3.72} &  \num{5.85} \\
$G$                          &                         \si{\milli\electronvolt} &   \num{4.23} &           \num{5.23} &  \num{4.89} &  \num{5.97} \\
$c_\text{v}$ \vspace{1pt}    &  \si[per-mode=fraction]{\cal\per\mol\per\kelvin} &  \num{0.0137} &           \num{0.0203} &  \num{0.0158} &   \num{0.0177}\\
\bottomrule
\end{tabular}
\end{adjustbox}
\vspace{1mm}
\caption{The table presents the MAE scores achieved by the \texttt{MMF W/GPT} models when predicting molecular properties on the QM9 dataset. All experiments were performed with the scaffold technique, using a value of $K$=16. In this context, $K$ denotes the number of training samples utilized in few-shot prompting.}
\label{tb:main_10}
\vspace{-3mm}
\end{table}

\vspace{-6mm}
\subsection{Hyperparameter Tuning}
\vspace{-2mm}
Hyperparameters are parameters that are not learned from data but are set prior to the training process. They have a direct impact on the performance of the framework. To optimize the performance of our \texttt{MMF} framework, we conducted in-depth hyperparameter tuning through careful experimentation and analysis. We chose to use random search as an efficient method for exploring hyperparameters and identifying the best framework configuration on benchmark datasets, rather than employing computationally intensive methods like grid search or Bayesian optimization. This approach allowed us to achieve optimal performance on the test dataset across various benchmark datasets, as measured by the MAE metric. We identified a set of hyperparameters that significantly improved the \texttt{MMF} framework's performance. We did not fine-tune large language models (LLMs) for the downstream property prediction task; instead, we accessed LLMs through text-based API interactions. We fine-tuned small-scale LMs using LLM-generated textual descriptions for the molecular property prediction task, while minimizing supervised regression loss. Hyperparameter optimization was performed on the \texttt{MMF-W/GPT-4} variant of our framework. The key prioritized hyperparameters for this framework are batch size ($b \in {32, 48, 64}$) and hidden or embedding dimension ($d \in {64, 128, 196, 256}$). Tables \ref{tb:main_11} and \ref{tb:main_12} present the results of hyperparameter tuning on representative benchmark datasets. We report the results for the near-optimal combinations of hyperparameters. All experiments were conducted using the GPT-4 framework with scaffold technique, setting $K$=16. In summary, based on the experimental results presented in tables \ref{tb:main_11} and \ref{tb:main_12}, the best hyperparameters for the \texttt{MMF} framework---particularly for the QM-8 dataset and many targets in the QM-9 dataset---appear to be a batch size ($b$) of 32 and an embedding dimension ($d$) of 128.

\vspace{-2mm}
\begin{table*}[h!]
\small
\begin{adjustbox}{width=0.7\columnwidth,center}
\centering
\def\arraystretch{1.1}
{%
\begin{tabular}{@{}c|cc@{}}
\toprule
Methods & Validation MAE ($\times 1.0e^{-3}$) & Test MAE ($\times 1.0e^{-3}$) \\
\midrule
MMF ($b$=32, $d$=128) &  \textbf{7.63 $\pm$ 0.07} & \textbf{7.45$\pm$ 0.03} \\
MMF ($b$=32, $d$=256) &  \text{8.34 $\pm$ 0.02} & \text{8.41$\pm$ 0.05}\\
MMF ($b$=48, $d$=128) &  \text{8.13 $\pm$ 0.06} & \text{8.25$\pm$ 0.03} \\
MMF ($b$=64, $d$=256) &  \text{8.57 $\pm$ 0.09} & \text{8.67$\pm$ 0.04} \\
\bottomrule
\end{tabular}
}
\end{adjustbox}
\vspace{-1mm}
\caption{The table presents the hyperparameter study results on the QM-8 dataset.}
\label{tb:main_11}
\vspace{-2mm}
\end{table*}

\vspace{-3mm}
\begin{table}[h]
\footnotesize
\begin{adjustbox}{width=0.975\columnwidth,center}
\begin{tabular}{@{}c|c|ccccc@{}}
\toprule
Target &                                             Unit &                      MMF($b$=32, $d$=128) &      MMF($b$=32, $d$=256) &  MMF($b$=48, $d$=128)  & MMF($b$=64, $d$=256) & \\
\hline
$\mu$                        &                                \si{D}       &             \num{0.0106} &  \num{0.0127}   &  \num{0.0119} & \num{0.0134} \\
$\alpha$                     &                               \si{\bohr^3}       &             \num{0.0219} &  \num{0.0277}  &  \num{0.0224} & \num{0.0311}\\
$\epsilon_\text{HOMO}$       &                         \si{\milli\electronvolt} &                 \num{18.43} &    \num{22.36}  &  \num{20.53} & \num{23.68}\\
$\epsilon_\text{LUMO}$       &                         \si{\milli\electronvolt} &                 \num{9.57} &    \num{10.78} &  \num{10.41} & \num{11.73} \\
$\Delta\epsilon$             &                         \si{\milli\electronvolt} &                   \num{22.34} &    \num{26.62} &  \num{24.05} &  \num{28.15}\\
$\left< R^2 \right>$         &                                     \si{\bohr^2} &    \textbf{\num{0.0108}} &   \textbf{\num{0.0121}}  &  \textbf{\num{0.0114}} & \textbf{\num{0.0139}}\\
ZPVE                         &                         \si{\milli\electronvolt} &                \num{0.785} &    \num{0.917} &  \num{0.842} & \num{0.989}\\
$U_0$                        &                         \si{\milli\electronvolt} &                 \num{3.55} &    \num{3.83}  &  \num{3.67} & \num{4.07}\\
$U$                          &                         \si{\milli\electronvolt} &                 \num{2.43} &    \num{2.55} &  \num{2.51} & \num{2.66}\\
$H$                          &                         \si{\milli\electronvolt} &                 \num{3.09} &    \num{3.31} &  \num{3.14} & \num{3.44} \\
$G$                          &                         \si{\milli\electronvolt} &                 \num{4.23} &    \num{4.68}  &  \num{4.41} & \num{4.97}\\
$c_\text{v}$ \vspace{1pt}    &  \si[per-mode=fraction]{\cal\per\mol\per\kelvin} &             \num{0.033} &   \num{0.081} &   \num{0.047} &  \num{0.099}\\
\bottomrule
\end{tabular}
\end{adjustbox}
\vspace{1mm}
\caption{The table presents the hyperparameter study results on the QM-9 dataset.}
\label{tb:main_12}
\vspace{-5mm}
\end{table}

\vspace{-4mm}
\subsection{Large Language Models and Prompting} 
\vspace{-2mm}
Large Language Models (LLMs) have brought about a paradigm shift in natural language processing for task adaptation with the `pre-train, prompt, and predict' approach. This approach allows for more generalized adaptation to various tasks by using natural language instructions as prompts, without the need for extensive fine-tuning. It is flexible and efficient, enabling LLMs to perform a wide range of NLP tasks with minimal task-specific adaptation. This is achieved through demonstrations that allow LLMs to learn from analogy. Consequently, it has replaced the previously established `pre-train, fine-tune' approach, which involves customizing the LLMs for each task through fine-tuning using task-specific labeled data. In this new paradigm, the LLM undergoes an initial phase of pre-training on a vast collection of unannotated text corpora without explicit human supervision. This process facilitates the learning of grammar, syntax, semantics, and even some level of common-sense reasoning, allowing the LLM to achieve improved linguistic comprehension and generate human-like responses. Instead of fine-tuning the language model with task-specific labeled data, as is customary in conventional approaches, this approach prompts the LLM with a natural language query that explicitly outlines the task and context. Based on this prompt, the LLM directly generates the desired output, using its pre-existing knowledge\cite{liu2023pre} embedded within its trainable parameters acquired during pretraining. A prompt is a textual input provided to an instruction-based or prompt-based language model to guide its behavior and generate desired outputs. It can take the form of a phrase, question, or sentence and may involve supplementary information or constraints. The prompt sets the context, provides instructions, and helps shape the LLM's responses to align with the desired outcome. LLMs possess the ability to generate responses based on contextual information and prior training, allowing them to provide context-aware and coherent responses in conversations. Moreover, LLMs can be further customized and enhanced through fine-tuning, which involves using reinforcement learning techniques\cite{ouyang2022training, sanh2021multitask} with human feedback to optimize their performance for specialized tasks or domains. This adaptability and versatility make LLMs powerful tools for a wide range of natural language processing applications. Consider an LLM, which accepts an input sequence of tokens $x=(x_1,x_2,...,x_n)$ and outputs a token sequence $y=(y_1,y_2,...,y_m)$. The LLM model is generally trained to optimize a conditional probability distribution $p(y|x)$, which assigns the probability to each possible output sequence $y$ given $x$. A prompt, denoted as $p$, can be integrated with the input sequence $x$ by concatenating the elements of $p$ to the beginning of the input sequence, resulting in a new sequence $\hat{x}=(p,x_1,x_2,...,x_n)$. We utilize $\hat{x}$ to calculate the conditional probability distribution $p(y|\hat{x})$. Formally, the probability of the output sequence $y$ given $\hat{x}$ can be represented as:

\vspace{-4mm}
\resizebox{1.0\linewidth}{!}{
\hspace{0mm}\begin{minipage}{\linewidth}
\begin{align}
 p(y|\hat{x})=\prod_{i=1}^m p(y_i|y_{<i}, \hat{x}) \nonumber
\end{align}
\end{minipage}
} 

\vspace{-3mm}
where $y_{<i}$ denotes the prefix of the sequence $y$ up to position $i-1$, and $p(y_i | y_{<i}, \hat{x})$ represents the probability of generating token $y_i$ given $y_{<i}$ and $\hat{x}$. The model's attention mechanism can identify the most relevant parts of the input prompt during each output token generation, facilitating the computation of $p(y_i | y_{<i}, \hat{x})$. In summary, the prompt plays a crucial role in shaping the conditional probability distribution, as it provides relevant information and context to guide the generation of the desired output. Zero-shot prompting is a process in NLP in which large language models (LLMs), such as GPT-3.5 or Google Bard, are equipped to perform specific tasks without undergoing task-specific training. This ability is derived from their initial training on extensive and diverse text corpora, enabling them to access a broad spectrum of general knowledge. This method utilizes detailed, context-rich prompts to outline the task at hand, guiding the LLM to generate suitable responses based on its foundational knowledge, thereby demonstrating the language model's zero-shot learning capabilities. Few-shot prompting is a technique in NLP that allows LLMs to perform a task with a few task-specific demonstrations. This is accomplished by first pre-training the LLM on a massive volume of text corpora, which allows it to learn a vast range of general knowledge. Then, the LLM is given a few task-specific input-output pairs and is trained to generate similar outputs. The LLM uses its pre-existing knowledge to generalize from the few demonstrations and learn to perform the task. In our work, we harness the zero-shot reasoning of LLMs, acquired through pre-training on vast text corpora. This empowers LLMs to enhance molecular graph-level embeddings from GNNs with richer textual information about molecular properties, reactivity, and more. This capability enables LLMs to tackle complex problems without requiring dedicated task-specific training. Table \ref{tab:lmprompts} shows the LLM(GPT-3.5)-retrieved text for a natural language query about an organic molecule in SMILES notation ``CC(=O)C". Additionally, we leverage LLMs' few-shot learning abilities to predict molecular properties of unseen molecules using demonstrations from training data that consists of chemical SMILES strings and property pairs. 

\bibliographystyle{plain}
\bibliography{reference}

\newpage
 
\begin{tabularx}{1.25\textwidth}{bss}
\caption{LLM(GPT-3.5)-retrieved text for a natural language query for an organic molecule in SMILES notation "CC(=O)C".}\label{tab:lmprompts}\\
\hline
\toprule
\textbf{Query prompt}  & \textbf{LLM-retrieved text} \\
\midrule
\endhead
\textcolor{black}{What is the molecular structure of this organic molecule in SMILES notation "CC(=O)C". Could you describe its atoms, bonds, functional groups, and overall arrangement?} & The organic molecule in SMILES notation "CC(=O)C" represents the compound acetone, which is a simple ketone. Let's break down the SMILES notation and describe its molecular structure:

\vspace{2mm}
SMILES notation: "CC(=O)C"

\vspace{2mm}
Atoms:

C: Represents carbon atoms.
O: Represents an oxygen atom.

\vspace{2mm}
Bonds:

Single bonds (represented by just the atom symbols): There are three single bonds between the carbon atoms (C-C).
Double bond (represented by "="): There is one double bond between the carbon and oxygen atoms (C=O).

\vspace{2mm}
Functional groups:

Carbonyl group: The functional group "C=O" is a carbonyl group, which is a carbon atom double-bonded to an oxygen atom.

\vspace{2mm}
Overall arrangement:
Acetone (CC(=O)C) is a simple three-carbon compound. It consists of three carbon atoms and one oxygen atom. The carbon atoms are connected in a chain, and one of the carbons is double-bonded to an oxygen atom, forming the carbonyl group. The molecule forms a linear structure with the carbonyl group in the middle. The "CH3" groups on both sides represent methyl (CH3) groups, which are single-bonded to the adjacent carbon atoms. The central carbon is double-bonded to the oxygen atom, forming the carbonyl functional group (C=O). Acetone is a common organic solvent and is found in some household products like nail polish removers and paint thinners. It is also used in chemical and industrial processes. \\

\hline
\hline

\textcolor{black}{What are the physical properties of this molecule such as its boiling point, melting point, and density?} & The physical properties of acetone (CC(=O)C) are as follows:

\vspace{2mm}
Boiling Point: The boiling point of acetone is approximately 56.5°C (approximately 133.7°F). This relatively low boiling point makes it a volatile liquid, which means it evaporates readily at room temperature.

\vspace{2mm}
Melting Point: The melting point of acetone is around -94.9°C (approximately -138.8°F). As with its boiling point, the low melting point is characteristic of many small organic molecules.

\vspace{2mm}
Density: The density of acetone is about 0.79 g/cm³ (at 20°C or 68°F). It is less dense than water, which has a density of 1 g/cm³ at the same temperature. This lower density is why acetone floats on water.

\vspace{2mm}
Acetone is a colorless, flammable liquid with a characteristic odor. It is highly miscible with water and many other organic solvents, making it a versatile solvent in various industries, laboratories, and domestic applications. It is important to note that these physical properties can vary slightly depending on the specific experimental conditions and purity of the acetone sample. \\

\hline
\hline

\textcolor{black}{What is the solubility behavior of this molecule? In which solvents does it dissolve and which does it not?} & Acetone (CC(=O)C) is a highly versatile solvent due to its unique solubility behavior. It is classified as a polar aprotic solvent, which means it has a significant dipole moment and can dissolve a wide range of polar and nonpolar substances. Here's a summary of its solubility behavior:

\vspace{2mm}
Solubility of Acetone:

Water (H2O): Acetone is highly miscible with water. It can dissolve in water in all proportions, forming a homogeneous mixture. This property makes acetone an excellent solvent for substances that are water-soluble or partially water-soluble.

Polar Solvents: Acetone is miscible with many polar solvents, including alcohols, ethers, ketones, and some chlorinated solvents. It can dissolve various organic compounds that contain polar functional groups, such as hydroxyl (OH) and carbonyl (C=O) groups.

Nonpolar Solvents: Acetone can also dissolve in nonpolar solvents to some extent. However, its solubility in nonpolar solvents is not as high as in polar solvents. Acetone's ability to dissolve nonpolar substances is due to its relatively small size and nonpolar "tail" (methyl groups) in its molecular structure.

\vspace{2mm}
Examples of Solvents in which Acetone Dissolves:

Ethanol,
Diethyl ether,
Ethyl acetate,
Chloroform,
Carbon tetrachloride,
Hexane.

\vspace{2mm}
Solvents in which Acetone Does Not Dissolve:
Acetone is not suitable as a solvent for some highly nonpolar substances or compounds that lack any polar functional groups. These may include certain hydrocarbons and nonpolar solvents like benzene and toluene.

\vspace{2mm}
As a general rule, acetone is an effective solvent for a wide range of organic compounds, but its solubility behavior depends on the specific properties of the substances being dissolved. Additionally, in some cases, the solubility of acetone may be limited by competing chemical reactions or phase separations, especially at high concentrations or with certain reactants. Always perform solubility tests and consult reference sources when choosing a solvent for a specific application.\\

\hline
\hline

\textcolor{black}{What is the chemical reactivity of this molecule? How does it interact with various reagents? } &  Acetone (CC(=O)C) is a chemically reactive molecule due to the presence of the carbonyl group (C=O) in its structure. Its chemical reactivity allows it to participate in various types of reactions with different reagents. Here are some common reactions that acetone can undergo:

\vspace{2mm}
Nucleophilic Addition Reactions: The carbonyl carbon in acetone is electrophilic, making it susceptible to attack by nucleophiles. In nucleophilic addition reactions, the C=O bond is broken, and a nucleophile adds to the carbonyl carbon. This leads to the formation of new compounds.

Example:
Acetone + Hydrogen Cyanide $\rightarrow$ Cyanohydrin 

\vspace{2mm}
Aldol Condensation: Acetone can undergo aldol condensation reactions, where two molecules of acetone react in the presence of a base to form a $\beta$-hydroxy ketone. This reaction is important in organic synthesis to build larger molecules. 

Example:
Acetone $\rightarrow$ Diacetone Alcohol

\vspace{2mm}
Reduction: Acetone can be reduced to produce different products, depending on the reagent used. For example, with sodium borohydride (NaBH4), acetone is reduced to yield isopropanol (2-propanol), while with lithium aluminum hydride (LiAlH4), it can be reduced to isopropyl alcohol (propan-2-ol).

Example:
Acetone + NaBH4 $\rightarrow$ Isopropanol

\vspace{2mm}
Oxidation: Acetone can be oxidized to form various products, including carboxylic acids or esters, depending on the oxidizing agent and reaction conditions. 

Example:
Acetone + Acidified Potassium Dichromate $\rightarrow$ Acetic Acid

\vspace{2mm}
Halogenation: Acetone can undergo halogenation reactions, where halogens (e.g., chlorine or bromine) add to the carbonyl carbon, resulting in halogenated derivatives.

Example:
Acetone + Bromine $\rightarrow$ 2,2-Dibromo-2-propanol

\vspace{2mm}
Polymerization: Acetone can polymerize under certain conditions, especially when exposed to strong acids or bases. This can lead to the formation of polyacetals.

\vspace{2mm}
Acetone's reactivity is crucial in various industrial processes and organic synthesis. However, its chemical reactivity also means that it should be handled with care and caution, especially when using it as a solvent or in chemical reactions, to prevent undesired reactions and ensure safety. \\

\hline
\hline

\textcolor{black}{Are there any common reactions that this molecule is known to undergo? Could you describe them?} & Acetone (CC(=O)C) is known to undergo several common reactions due to the presence of its carbonyl group (C=O). Some of the most important reactions involving acetone are:

\vspace{2mm}
Nucleophilic Addition Reactions:
Acetone can undergo nucleophilic addition reactions, where a nucleophile attacks the electrophilic carbon of the carbonyl group. This results in the breaking of the C=O bond and the formation of a new compound. One common nucleophilic addition reaction involving acetone is the formation of cyanohydrins:
Example:
Acetone + Hydrogen Cyanide $\rightarrow$ Cyanohydrin

\vspace{2mm}
Aldol Condensation:
Acetone can undergo aldol condensation, a reaction where two molecules of acetone or an acetone molecule and another aldehyde or ketone react in the presence of a base. This reaction leads to the formation of a $\beta$-hydroxy ketone or aldol compound.
Example:
Acetone → Diacetone Alcohol

\vspace{2mm}
Haloform Reaction:
Acetone can undergo the haloform reaction in the presence of halogens (e.g., chlorine or bromine) and hydroxide ions. This reaction results in the formation of a halogenated methyl ketone and a halide ion.
Example:
Acetone + Chlorine + Hydroxide $\rightarrow$ Chloroform + Acetate Ion

\vspace{2mm}
Cannizzaro Reaction:
Under strong basic conditions, acetone can undergo the Cannizzaro reaction, which involves the disproportionation of the molecule into an alcohol and a carboxylic acid.
Example:
Acetone + Hydroxide $\rightarrow$ Isopropanol + Acetate Ion

\vspace{2mm}
Dehydration:
Acetone can undergo dehydration in the presence of acid catalysts, such as sulfuric acid. In this reaction, water is eliminated from acetone to form an unsaturated compound.
Example:
Acetone $\rightarrow$ 2-Butanone (Methyl Ethyl Ketone)

\vspace{2mm}
Polymerization:
Acetone can undergo polymerization under certain conditions, leading to the formation of polyacetals. These are just a few examples of the common reactions involving acetone. As a versatile solvent and reagent, acetone participates in various chemical processes in both industrial and laboratory settings. Its reactivity and ability to form a wide range of products make it an essential component in organic synthesis and many chemical applications.\\

\hline
\hline

\textcolor{black}{Does this molecule exhibit any unique optical, electrical, or magnetic properties?} & Acetone (CC(=O)C) is a simple organic molecule that does not exhibit any unique optical, electrical, or magnetic properties under standard conditions. It is a colorless liquid at room temperature and does not display any inherent color, making it optically inactive. Acetone is also a non-conductive material, meaning it is not an electrical conductor, and it does not possess any special electrical properties.

\vspace{2mm}
Regarding magnetism, acetone is a diamagnetic substance. Diamagnetic materials are those that generate a weak magnetic field in opposition to an externally applied magnetic field. However, the diamagnetic effect is very weak in acetone and most organic compounds. As a result, the magnetic properties of acetone are generally not significant for most practical purposes.

\vspace{2mm}
It's worth noting that the lack of unique optical, electrical, or magnetic properties does not diminish the importance and versatility of acetone as a solvent and reactant in various chemical processes and applications. Its value lies in its role as a solvent, its ability to participate in various chemical reactions, and its use as an intermediate in the synthesis of various compounds in organic chemistry and industry. \\

\hline
\hline

\textcolor{black}{Is this molecule chiral? If yes, how does its chirality influence its behavior or properties?} & The acetone molecule (CC(=O)C) is not chiral. Chirality is a property of molecules that have non-superimposable mirror images, known as enantiomers. Chiral molecules possess an asymmetric carbon atom (a carbon atom bonded to four different groups), and this asymmetry gives rise to the existence of two enantiomeric forms. 

\vspace{2mm}
In acetone, all three carbon atoms are bonded to two hydrogen atoms and one methyl (CH3) group each. Since all the carbon atoms have the same groups bonded to them, there is no chiral center, and acetone does not have enantiomers.The absence of chirality in acetone means that it behaves as an achiral molecule. It exhibits typical chemical and physical properties of a small ketone, such as its ability to act as a solvent, participate in various chemical reactions, and display its characteristic odor and volatility. 

\vspace{2mm}
Chirality can play a significant role in the behavior and properties of some molecules, especially in biological systems. Chiral molecules often interact differently with other chiral molecules or biomolecules, such as enzymes or receptors, leading to different biological activities. The two enantiomers of a chiral molecule can have different pharmacological effects, toxicities, and physiological behaviors, which is important in drug development and medicinal chemistry. 

\vspace{2mm}
However, since acetone is achiral, it does not exhibit such differences between enantiomers. Its chemical and physical properties are consistent, and both enantiomeric forms would be identical if they were to exist, which, in this case, they do not.  \\

\hline
\hline

\textcolor{black}{Is this molecule synthesized industrially or in the laboratory? If yes, could you explain the process?} & Acetone (CC(=O)C) is synthesized both industrially and in the laboratory. It is produced on a large scale industrially due to its wide range of applications as a solvent and chemical intermediate. The primary industrial method for acetone production is through the cumene process, which involves the following steps:

\vspace{2mm}
Industrial Synthesis of Acetone via Cumene Process:

\vspace{2mm}
Step 1: Propylene and Benzene Alkylation:
Propylene and benzene are reacted in the presence of an acid catalyst (typically aluminum chloride, AlCl3) to produce cumene (isopropylbenzene) through alkylation.

CH3CH=CH2 + C6H6 $\rightarrow$ (CH3)2C6H5 (cumene) 

\vspace{2mm}
Step 2: Cumene Oxidation:
Cumene is then oxidized using air (oxygen) in the presence of a metal catalyst (usually cobalt or manganese acetate) to form cumene hydroperoxide.

(CH3)2C6H5 + O2 $\rightarrow$ (CH3)2C6H5OOH (cumene hydroperoxide)

\vspace{2mm}
Step 3: Cleavage of Cumene Hydroperoxide:
Cumene hydroperoxide is further cleaved into phenol and acetone using an acid catalyst (sulfuric acid).

(CH3)2C6H5OOH $\rightarrow$ C6H5OH (phenol) + (CH3)2CO (acetone)

The acetone obtained from the cumene process is typically purified through various distillation steps to obtain a high-purity product suitable for various industrial applications. 

\vspace{2mm}
Laboratory Synthesis of Acetone:

In the laboratory, acetone can be synthesized through various methods, including the hydration of propyne (also known as methylacetylene or propargylene) or the dehydrogenation of isopropanol (2-propanol). However, these methods are less common than the industrial cumene process due to cost and efficiency considerations.

\vspace{2mm}
For example, the hydration of propyne involves reacting propyne with water in the presence of a strong acid catalyst to form acetone:

H\chemfig{C ~ C}-CH3 + H2O $\rightarrow$ (CH3)2CO 

\vspace{2mm}
The dehydrogenation of isopropanol can also produce acetone:

(CH3)2CHOH $\rightarrow$ (CH3)2CO + H2

\vspace{2mm}
In the laboratory, acetone is often synthesized as needed for specific experiments or applications rather than produced on an industrial scale. The choice of the synthesis method depends on factors such as availability, cost, and the desired purity of the acetone product. \\

\textcolor{black}{Are there any notable uses or applications for this molecule in medicine, industry, or other fields?} & Acetone (CC(=O)C) has numerous notable uses and applications in various fields, including medicine, industry, and other areas. Some of the key applications are as follows:

\vspace{2mm}
Solvent:
Acetone is a versatile and widely used solvent in various industries, including pharmaceuticals, paints, coatings, adhesives, and plastics. It is an effective solvent for a wide range of organic compounds and is often used for cleaning and degreasing purposes.

\vspace{2mm}
Chemical Intermediate:
Acetone serves as an important chemical intermediate in the production of various compounds. It is a precursor for the synthesis of chemicals such as methyl methacrylate (used in the production of acrylic plastics), bisphenol-A (used in the production of polycarbonate plastics), and mesityl oxide (used as a solvent and in perfumes).

\vspace{2mm}
Cleaning Silicon Wafers:
Acetone is employed in the electronics industry to clean and remove contaminants from silicon wafers used in semiconductor manufacturing.

\vspace{2mm}
Pharmaceuticals:
In the pharmaceutical industry, acetone is used in the synthesis and formulation of various drugs and medications. It is employed to dissolve and stabilize active pharmaceutical ingredients and excipients.

\vspace{2mm}
Laboratory Reagent:
Acetone is frequently used in laboratory settings as a reagent for various chemical reactions and as a solvent for cleaning glassware and equipment.

\vspace{2mm}
Cosmetic and Personal Care Products:
Acetone is used in cosmetic and personal care products such as lotions, creams, and astringents, where it acts as a solvent and helps deliver active ingredients into the skin.

\vspace{2mm}
Extractive Distillation:
In certain chemical processes, acetone is used as an extractive distillation agent to separate mixtures and purify various compounds.

\vspace{2mm}
Fuel Additive:
Acetone has been studied as a potential fuel additive that may enhance the combustion efficiency of certain fuels, although its practical use in this regard is limited.

\vspace{2mm}
It is essential to handle acetone with care, as it is a flammable liquid and can pose health risks if not used or stored properly. Its versatility, as well as its physical and chemical properties, make acetone a valuable and widely employed compound in various industries and applications.  \\

\hline
\hline

\textcolor{black}{Is this molecule found naturally? If yes, in what sources is it most commonly found? } &  Acetone (CC(=O)C) can be found naturally in trace amounts in certain sources. However, its natural occurrence is generally limited, and it is not a prominent component in the natural environment. Acetone is primarily produced through biological processes, and its concentration in natural sources is typically quite low. Some of the common natural sources of acetone include:

\vspace{2mm}
Human and Animal Metabolism:
Small amounts of acetone are produced as a byproduct of metabolism in humans and animals. It is formed as a result of the breakdown of fatty acids, amino acids, and carbohydrates in the body.

\vspace{2mm}
Plants:
Certain plants are capable of producing acetone as part of their metabolism. Acetone can be found in trace amounts in some plant tissues.

\vspace{2mm}
Volcanic Emissions:
Acetone has been detected in volcanic emissions and geothermal areas, likely as a result of microbial activity in the volcanic environment.

\vspace{2mm}
Forest Fires:
Acetone can be present in the smoke and emissions from forest fires due to the combustion of organic materials. 

\vspace{2mm}
While acetone occurs naturally in these sources, its concentrations are generally very low, and it is typically overshadowed by other more abundant volatile organic compounds (VOCs). In most cases, the levels of naturally occurring acetone are not significant enough to have a major impact on the environment or human health. The primary sources of acetone in the environment are industrial processes, which release much higher quantities of acetone into the air and water. The industrial production and use of acetone for various applications contribute significantly to its presence in the environment. \\

\hline
\hline

\textcolor{black}{What safety measures should be taken when handling this molecule? } &  Handling acetone (CC(=O)C) requires some safety precautions due to its flammable and volatile nature. Here are important safety measures to consider when working with acetone:

\vspace{2mm}
Flammability:
Acetone is highly flammable and can form explosive vapor-air mixtures. Keep acetone away from open flames, sparks, and sources of heat. Use it only in well-ventilated areas, and avoid using it near ignition sources.

\vspace{2mm}
Proper Storage:
Store acetone in tightly sealed, non-reactive containers away from direct sunlight, heat, and ignition sources. Ensure that the storage area is well-ventilated and separate from incompatible chemicals.

\vspace{2mm}
Personal Protective Equipment (PPE):
When handling acetone, wear appropriate personal protective equipment, including chemical-resistant gloves, safety goggles or a face shield, and a lab coat or protective clothing to prevent contact with skin and eyes.

\vspace{2mm}
Ventilation:
Work with acetone in a well-ventilated area, such as a fume hood, to reduce the risk of inhalation exposure to vapors.

\vspace{2mm}
Avoid Skin Contact:
Acetone can cause skin dryness and irritation. Minimize skin contact, and promptly wash off any spills on the skin with soap and water. If skin irritation occurs, seek medical attention. 

\vspace{2mm}
Avoid Inhalation:
Acetone vapors can be irritating to the respiratory tract. Use acetone in a well-ventilated area or wear a suitable respirator if exposure levels are high.

\vspace{2mm}
No Smoking Policy:
Do not smoke while handling acetone, and enforce a strict no-smoking policy in areas where acetone is used or stored.

\vspace{2mm}
Fire Safety:
Keep fire extinguishing equipment, such as fire extinguishers, nearby in case of accidental fires. Know the appropriate fire suppression methods for flammable liquid fires.

\vspace{2mm}
Spill Response:
In the event of a spill, contain and absorb the acetone with suitable absorbent materials, and follow proper disposal procedures for hazardous waste.

\vspace{2mm}
Labeling and Storage Compatibility:
Label containers containing acetone with appropriate hazard warnings and identification. Store acetone away from incompatible chemicals to prevent accidental reactions.

\vspace{2mm}
Disposal:
Dispose of acetone waste in accordance with local, state, and federal regulations. Do not dispose of acetone down the drain or in regular trash.

\vspace{2mm}
It is important to review and follow the safety data sheet (SDS) provided by the manufacturer for specific handling instructions and safety precautions. When using acetone in an industrial or laboratory setting, ensure that personnel are trained in proper handling procedures and emergency response protocols. \\

\hline
\hline

\textcolor{black}{Are there any environmental impacts associated with the production, use, or disposal of this molecule?} &  There are potential environmental impacts associated with the production, use, and disposal of acetone (CC(=O)C). While acetone is a naturally occurring compound in small amounts, the significant quantities produced and used in various industries can contribute to environmental concerns. Some of the environmental impacts include:

\vspace{2mm}
Air Pollution:
Acetone is a volatile organic compound (VOC) and is emitted into the atmosphere during its production, use as a solvent, and various industrial processes. VOCs can contribute to the formation of ground-level ozone and smog, which can have negative effects on air quality and human health.

\vspace{2mm}
Water Contamination:
Improper disposal of acetone or industrial wastewater containing acetone can lead to its release into water bodies. Acetone is relatively soluble in water, and its presence in water sources can have adverse effects on aquatic ecosystems and wildlife.

\vspace{2mm}
Soil Contamination:
Accidental spills or improper disposal of acetone-containing waste can lead to soil contamination. Acetone is capable of migrating through soil and potentially contaminating groundwater.

\vspace{2mm}
Non-Target Species:
Acetone's volatility and widespread use may result in its presence in various environmental compartments. Although acetone is not considered highly toxic to humans, it can be harmful to certain aquatic organisms and other non-target species if present in elevated concentrations.

\vspace{2mm}
Ozone Depletion:
While acetone itself is not an ozone-depleting substance, its industrial production and use may involve processes that generate other ozone-depleting substances or greenhouse gases. The release of these substances can contribute to stratospheric ozone depletion and climate change.

\vspace{2mm}
To mitigate these environmental impacts, it is essential to adopt proper waste management practices, control emissions during production and industrial processes, and promote the responsible use and handling of acetone. Implementing technologies for VOC capture and recovery can help reduce air emissions. Recycling and reusing acetone waste can also be beneficial in reducing its environmental impact. Additionally, regulatory measures and best management practices can be implemented to ensure the safe and sustainable use of acetone in various industries. \\

\bottomrule
\end{tabularx}


\end{document}